\documentclass[pdflatex,sn-mathphys-num]{sn-jnl}

\UseRawInputEncoding
\usepackage{float}  
\usepackage[T1]{fontenc}
\usepackage{lmodern}  

\usepackage{newtxtext}
\usepackage{graphicx}%
\usepackage{multirow}%
\usepackage{amsmath,amssymb,amsfonts}%
\usepackage{amsthm}%
\usepackage{mathrsfs}%
\usepackage[title]{appendix}%
\usepackage{xcolor}%
\usepackage{textcomp}%
\usepackage{manyfoot}%
\usepackage{booktabs}%
\usepackage{algorithm}%
\usepackage{algorithmicx}%
\usepackage{algpseudocode}%
\usepackage{listings}%
\usepackage{subcaption}
\usepackage[normalem]{ulem}
\usepackage{comment}

\makeatletter
\renewcommand\paragraph{\@startsection{paragraph}{4}{0pt}%
   {3.25ex \@plus1ex \@minus.2ex}%
   {-1em}%
   {\normalfont\normalsize\bfseries}} 
\makeatother

\theoremstyle{thmstyleone}%
\theoremstyle{thmstyletwo}%

\theoremstyle{thmstylethree}%

\begin{document}

\title[Lang1]{Generalist Foundation Models Are Not Clinical Enough for Hospital Operations}

\author*[1,2,3]{\fnm{Lavender Y.} \sur{Jiang}}\email{Lavender.Jiang@nyu.edu}
\equalcont{These authors contributed equally to this work.}

\author[1]{\fnm{Angelica} \sur{Chen}}\email{Angelica.Chen@nyu.edu}
\equalcont{These authors contributed equally to this work.}

\author[2]{\fnm{Xu} \sur{Han}}\email{Xu.Han@nyulangone.org}

\author[2,4]{\fnm{Xujin Chris} \sur{Liu}}\email{xl3942@nyu.edu}

\author[1,2,3]{\fnm{Radhika} \sur{Dua}}\email{rd3571@nyu.edu }

\author[5,6]{\fnm{Kevin} \sur{Eaton}}\email{Kevin.Eaton@nyulangone.org}

\author[2,7]{\fnm{Frederick} \sur{Wolff}}\email{wfrederick@ethz.ch}

\author[2,8]{\fnm{Robert} \sur{Steele}}\email{Robert.Steele@nyulangone.org}

\author[9,10]{\fnm{Jeff} \sur{Zhang}}\email{Jeff.Zhang@nyulangone.org}

\author[2,11]{\fnm{Anton} \sur{Alyakin}}\email{Anton.Alyakin@nyulangone.org}

\author[2]{\fnm{Qingkai} \sur{Pan}}\email{ivrinom@gmail.com}

\author[2,12]{\fnm{Yanbing} \sur{Chen}}\email{yc6785@nyu.edu}

\author[2,5]{\fnm{Karl L.} \sur{Sangwon}}\email{ Karl.Sangwon@nyulangone.org}

\author[2,5]{\fnm{Daniel A.} \sur{Alber}}\email{Daniel.Alber@nyulangone.org}

\author[2]{\fnm{Jaden} \sur{Stryker}}\email{Jaden.Stryker@nyulangone.org}

\author[2,3,11]{\fnm{Jin Vivian} \sur{Lee}}\email{jin.v.lee@wustl.edu}

\author[6,9,10]{\fnm{Yindalon} \sur{Aphinyanaphongs}}\email{yin.a@nyulangone.org}

\author[1,3,13]{\fnm{Kyunghyun} \sur{Cho}}\email{Kyunghyun.Cho@nyu.edu}

\author*[1,2,3,5,9,14]{\fnm{Eric Karl} \sur{Oermann}}\email{Eric.Oermann@nyulangone.org}

\affil[1]{\orgdiv{Courant Institute School of Mathematics, Computing, and Data Science}, \orgname{New York University}, \orgaddress{\street{60 5th Ave}, \city{New York}, \postcode{10001}, \state{NY}, \country{USA}}}

\affil[2]{\orgdiv{Department of Neurosurgery}, \orgname{NYU Langone Health}, \orgaddress{\street{550 First Avenue}, \city{New York}, \postcode{10016}, \state{NY}, \country{USA}}}

\affil[3]{\orgdiv{Global AI Frontier Lab}, \orgname{New York University}, \orgaddress{\street{1 Metrotech Center, Fl. 22}, \city{Brooklyn}, \postcode{11201}, \state{NY}, \country{USA}}}

\affil[4]{\orgdiv{Electrical and Computer Engineering}, \orgname{Tandon School of Engineering}, \orgaddress{\street{6 MetroTech Center}, \city{Brooklyn}, \postcode{11201}, \state{NY}, \country{USA}}}

\affil[5]{\orgdiv{Grossman School of Medicine}, \orgname{New York University}, \orgaddress{\street{550 First Avenue}, \city{New York}, \postcode{10016}, \state{NY}, \country{USA}}}

\affil[6]{\orgdiv{Department of Medicine}, \orgname{NYU Langone Health}, \orgaddress{\street{550 First Avenue}, \city{New York}, \postcode{10019}, \state{NY}, \country{USA}}}

\affil[7]{\orgdiv{Department of Computer Science}, \orgname{ETH Zurich}, \orgaddress{\street{Universit\"atstrasse 6}, \city{City}, \postcode{8092}, \state{Zurich}, \country{Switzerland}}}

\affil[8]{\orgdiv{Department of Surgery}, \orgname{NYU Langone Health}, \orgaddress{\street{1  Park Avenue}, \city{New York}, \postcode{10016}, \state{NY}, \country{USA}}}

\affil[9]{\orgdiv{Division of Applied AI Technologies}, \orgname{NYU Langone Health}, \orgaddress{\street{227 East 30th Street}, \city{New York}, \postcode{10016}, \state{NY}, \country{USA}}}

\affil[10]{\orgdiv{Department of Population Health}, \orgname{NYU Langone Health}, \orgaddress{\street{450 First Avenue}, \city{New York}, \postcode{10019}, \state{NY}, \country{USA}}}

\affil[11]{\orgdiv{School of Medicine}, \orgname{Washington University of St. Louis}, \orgaddress{\street{660 S. Euclid Ave.}, \city{St. Louis}, \postcode{63110}, \state{MO}, \country{USA}}}

\affil[12]{\orgdiv{School of Global Public Health}, \orgname{New York University}, \orgaddress{\street{708 Broadway}, \city{New York}, \postcode{10003}, \state{NY}, \country{USA}}}

\affil[13]{\orgdiv{Prescient Design}, \orgname{Genentech}, \orgaddress{\street{149 5th Ave. 3rd floor}, \city{New York}, \postcode{10019}, \state{NY}, \country{USA}}}

\affil[14]{\orgdiv{Department of Radiology}, \orgname{NYU Langone Health}, \orgaddress{\street{450 First Avenue}, \city{New York City}, \postcode{10019}, \state{NY}, \country{USA}}}

\abstract{Hospitals and healthcare systems rely on operational decisions that determine patient flow, cost, and quality of care. Despite strong performance on medical knowledge and conversational benchmarks, foundation models trained on general text may lack the specialized knowledge required for these operational decisions. We introduce \textbf{\textsc{Lang1}}, a family of models (100M--7B parameters) pretrained on a specialized corpus blending 80 billion clinical tokens from NYU Langone Health's electronic health records (EHRs) and 627 billion tokens from the internet. To rigorously evaluate \textsc{Lang1} in real-world settings, we developed the REalistic Medical Evaluation (\textbf{\textsc{ReMedE}}), a benchmark derived from 668,331 EHR notes that evaluates five critical tasks: 30-day readmission prediction, 30-day mortality prediction, length of stay, comorbidity coding, and predicting insurance claims denial. 
In zero-shot settings, both general-purpose and specialized models underperform on four of five tasks (36.6\%--71.7\% AUROC), with mortality prediction being an exception (up to 94.2\% AUROC). 
After finetuning, \textsc{Lang1-1B} \textbf{outperforms} finetuned generalist models up to \unboldmath{$70\times$} larger and zero-shot models up to \unboldmath{$671\times$} larger, improving AUROC by 3.64\%--6.75\% and  1.66\%--23.66\% respectively. We also observed cross-task scaling with joint finetuning on multiple tasks leading to across the board improvement on other tasks. \textsc{Lang1-1B} effectively \textbf{transfers} to out-of-distribution settings, including other clinical tasks and an external health system. Our findings suggest that predictive capabilities for hospital operations \textbf{require} explicit supervised finetuning, and that this finetuning process is made more efficient by in-domain pretraining on EHR. Our findings support the emerging view that specialized LLMs can compete with generalist models in specialized tasks, and show that effective healthcare systems AI requires the combination of in-domain pretraining, supervised finetuning, and  real-world evaluation beyond proxy benchmarks.}


\maketitle
\section{Main}\label{sec:main}

Healthcare systems face high-stakes operational decisions daily: which patients are at imminent risk of decline, who can be safely discharged, how many beds will be available for new admissions. These decisions directly impact resource allocation, care coordination, and patient outcomes \citep{Sinsky2016-yo, Hingle2016-fx, Murphy2016-ym}. As healthcare systems face increasing patient volume, there is a need for tools that can analyze complex clinical data to inform these critical decisions. Foundation models, with their powerful text comprehension capabilities and versatility in specialized domains \citep{Yao2022-qs,hoffmann2022chinchilla, DeepSeek-AI_Guo_Yang_Zhang_Song_Zhang_Xu_Zhu_Ma_Wang_et_al._2025, Trinh_Wu_Le_He_Luong_2024, Gottweis_Weng_Daryin_Tu_Palepu_Sirkovic_Myaskovsky_Weissenberger_Rong_Tanno_et_al._2025, Wu_Irsoy_Lu_Dabravolski_Dredze_Gehrmann_Kambadur_Rosenberg_Mann_2023}, have emerged as a promising technology for optimizing hospital operations.

However, deploying LLMs in clinical settings is fraught with challenges. While LLMs show  promise in various clinical tasks~\citep{chen_2022_diaformer, chen_2023_dxformer, panagoulias_2023_evaluatingchatgpt, amin_2023_radiology_reports, ellershaw_2024_discharge, zhou_2024_diagnosis, zhang_2024_radiology_impressions, he_2024_lab_results, kweon_2024_ehrnoteqa, glicksberg_2024_admission, nazyrova_2024_readmission, shoham_2024_cpllm, scarlat_2025_mortality_readmission, bhasuran_2025_lab_results, McDuff2025-vc, alyakin_2025_neurosurgery_imaging}, there is disagreement on whether smaller specialized models ("specialists") can outperform general-purpose models ("generalists")~\citep{Nori2023CanGF, Zhou2024-td, Lehman2023-qn}. Many evaluations rely on proxy benchmarks that weakly reflect real-world clinical constraints like data scarcity and temporal shifts \citep{Johri_Jeong_Tran_Schlessinger_Wongvibulsin_Barnes_Zhou_Cai_Van_Allen_Kim_et_al._2025, Jiang2025-ei, Bean2025ClinicalKI, vishwanath2025medicallargelanguagemodels, Yan2025LLMSE, he_2024_lab_results, Hager2024-cz,Arora2025-us}. Data privacy concerns further limit the pretraining data for the vast majority of the clinical LLMs \citep{wornow_2023_shaky} to a small set of public corpora, MIMIC \citep{johnson_2023_mimic_iv} and PubMed \citep{pubmed.govDownloadPubMedData}, even though large-scale EHR datasets are known to improve out-of-domain generalization \citep{sushil_2022_clinical, yang_2022_gatortron, peng_2023_gatortrongpt, Jiang2023-ig, Lehman2023-qn}.

In this work, we focus on hospital operation tasks that represent the daily challenge of healthcare delivery and explore the tradeoffs between off-the-shelf generalists and specialized models trained on a health system's internal patient notes. Our contributions are as follows:

\begin{enumerate}

\item \textbf{\textsc{Lang1}: a family of models specialized for hospital operations.} We present \textsc{Lang1}, a suite of decoder LLMs (100M, 1B, and 7B), pretrained from scratch on a mix of 80 billion tokens of EHR notes and 627 billion tokens of internet texts. After task-specific finetuning, \textsc{Lang1} \textbf{outperforms} both off-the-shelf generalist LLMs (such as\textsc{Deepseek R1} 671B) and the parameter-efficient finetuned variant (LoRA finetuned \textsc{Deepseek Distilled Llama 70B}), on the \textsc{ReMedE} benchmark. Instruction finetuned on one or more tasks, \textsc{Lang1} \textbf{transfers} zero-shot to related tasks and to a different hospital, surpassing generalist models of similar scales. 

\item \textbf{\textsc{ReMedE}: an operations-grounded evaluation suite.} We construct an internal evaluation benchmark consisting of five clinically important classification tasks from a multi-hospital academic health system across 10 years, where each task contains 87,974 to 421,429 real patients. The benchmark has time-based splits to mimic deployment settings and data-efficiency protocols to reflect real-world constraints. 

\item \textbf{Engineering Principles for clinical utilities.} We analyze the training dynamics and show that pretraining on next token prediction of unlabeled clinical notes and web text contributes to emergent skills on comprehension tasks but is insufficient for excelling on \textsc{ReMedE}, which specifically requires supervised finetuning (SFT). However, SFT is made more efficient by in-domain pretraining, and larger models pretrained on more clinical data improve temporal robustness.

\end{enumerate}

Overall, our results suggest that health systems with the capacity for in-house model development can gain clear advantages from smaller specialized models, providing a practical and data-efficient pathway to robust operational prediction with minimal task-specific supervision.

\begin{figure}[ht!]
    \centering
    \includegraphics[width=\linewidth]{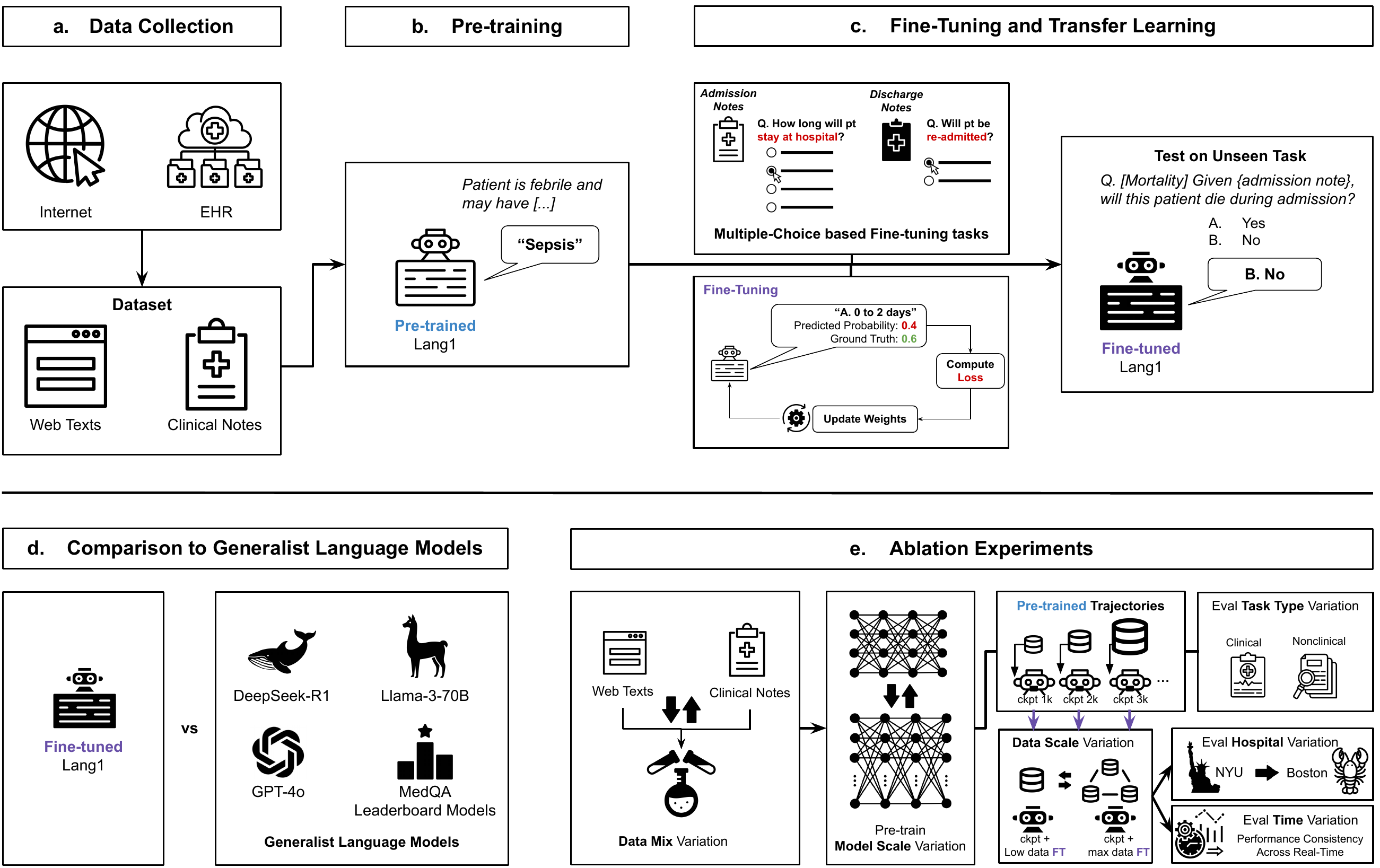}
    \caption{Overview. (a) We mix unlabeled EHR notes and web texts as our pretrain corpus. (b) We pretrain using next token prediction. (c) Instruction finetuning in multiple choice format enables cross-task transfer. (d) We compare Lang1 to off-the-shelf generalist models. (e) In order to derive design principles, we do ablations on data mix, model scale, pretrain trajectories, data scale, eval task type, eval hospital, and eval time.}
    \label{fig:overview}
\end{figure}

\subsection{Overview}

Our work consists of five stages: data collection, pretraining, finetuning, evaluation, and ablations.  As shown in \autoref{fig:overview}, we first collect unlabeled clinical notes from NYU Langone EHR and web texts from the internet and mix them to form the corpus (Figure~1a). We pretrain \textsc{Lang1} via next token prediction (Figure~1b). We instruction finetune \textsc{Lang1} to predict labels for real-world clinical tasks, enabling cross-task transfer (Figure~1c). We then compare finetuned \textsc{Lang1} with off-the-shelf generalist models (Figure~1d) and perform ablations of the data mix, model scale, eval task type, pretraining trajectory, finetune data scale, eval data time, and eval hospital to derive design principles for clinical utilities (Figure~1e).

We evaluate \textsc{Lang1} using \textsc{ReMedE}, an internal benchmark of real-world, high-impact clinical tasks beyond diagnosis. Unlike recent benchmarks that focus on multi-turn diagnostic dialogue \citep{zhou_2024_diagnosis, Johri_Jeong_Tran_Schlessinger_Wongvibulsin_Barnes_Zhou_Cai_Van_Allen_Kim_et_al._2025, McDuff2025-vc}, which captures an important but narrow part of clinical decision making, \textsc{ReMedE} is based on 668,331 EHR notes and emphasizes operational tasks that better represent the day-to-day challenges of healthcare delivery \citep{Sinsky2016-yo, Hingle2016-fx, Murphy2016-ym}. These tasks support practical goals such as reducing costs, optimizing resource use, and improving continuity of care. 

\textsc{ReMedE} includes five predictive tasks drawn from real-world hospital workflows: 30-day all-cause readmission, in-hospital mortality, binned length of stay, insurance denial, and imputation of binned Charlson Comorbidity Index (CCI, a measure of patient comorbidity burden). These tasks reflect critical decisions tied to patient outcomes, resource planning, and healthcare operations (See Methods \ref{sec:finetune_data} for more details). To assess model robustness to temporal distribution shifts, each task is evaluated across three non-overlapping test splits drawn from distinct time periods (\autoref{sec:timeline_viz}). \textsc{ReMedE} supports flexible few-shot evaluation across a range of language model interfaces and task formats. It is easily extensible for new tasks and evaluation settings. We plan to release \textsc{ReMedE} as a secure evaluation service, allowing trusted researchers to submit models and receive standardized evaluation results without direct access to patient data. This design safeguards patient privacy while enabling fair and reproducible model comparison. 

\section{Results}
\vspace{-10pt}

\begin{figure}[ht!]
    \centering
    \begin{subfigure}{\linewidth}
    \centering
         \includegraphics[width=.7\linewidth]{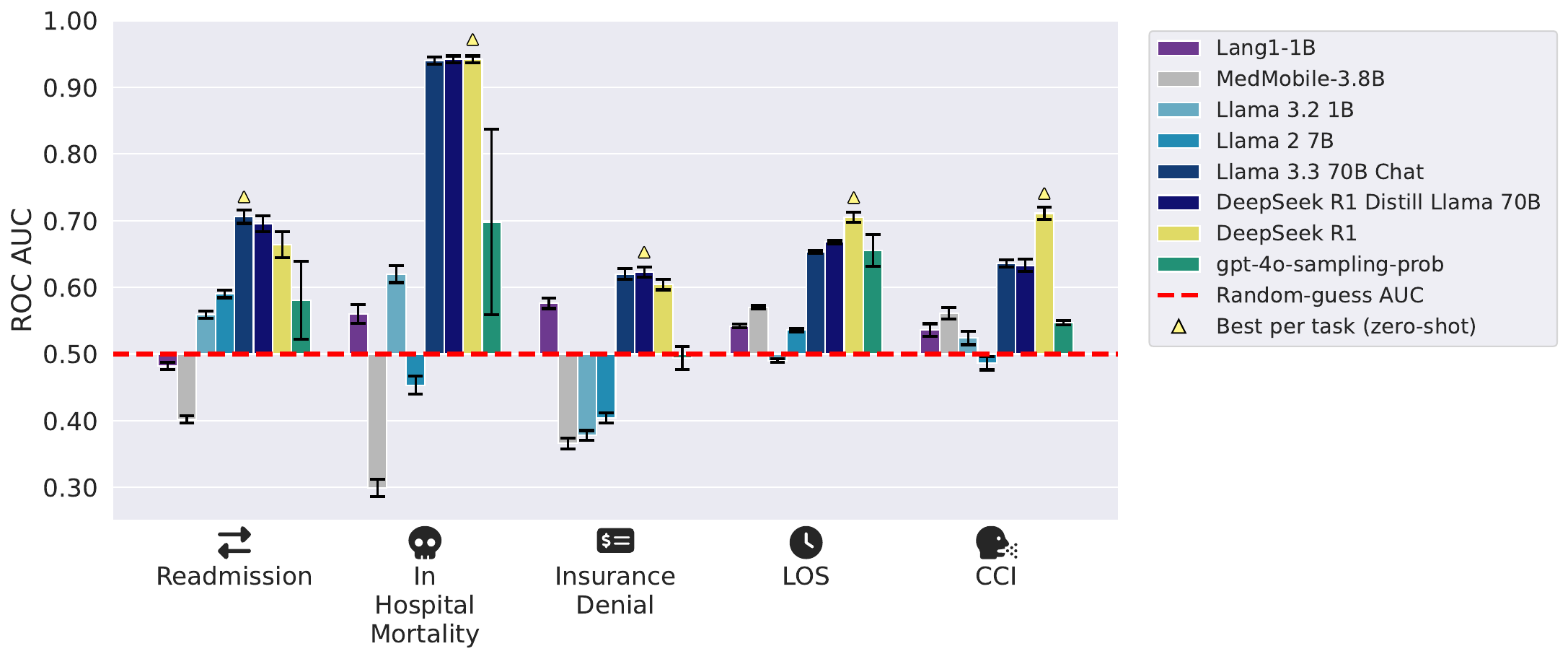}
    
         \caption{Both generalists and specialists underperform zero-shot. Yellow triangles indicate the best zero-shot performance per task. While the best mortality prediction AUROC is 94.2\%, performance of other tasks (readmission, insurance denial, LOS, CCI) range from 36.6\%--71.7\% AUROC.} 
         \label{fig:leaderboard-zero-shot}
    \end{subfigure}\hfill
    \begin{subfigure}{\linewidth}
    \centering
         \includegraphics[width=.7\linewidth]{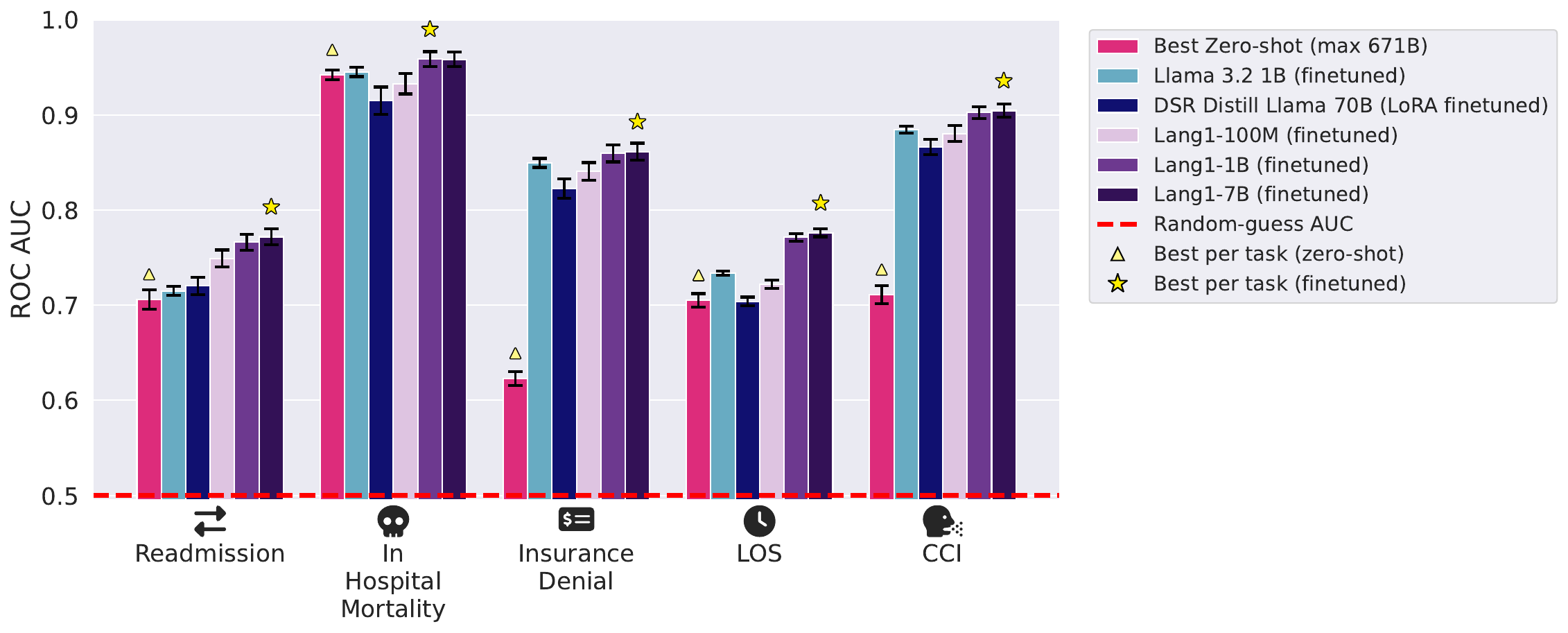}
         \caption{Finetuned \textsc{Lang1-1B} (purple) outperform best zero-shot performance (magenta) by 1.66\% to 23.66\% AUROC and finetuned \textsc{Llama 3.2 1B} (light blue) and LoRA finetuned \textsc{Llama 70B} (deep blue) by 3.64\% to 6.75\% AUROC. Yellow stars indicate the best finetuned performance per task.}
         \label{fig:leaderboard-finetuning}
    \end{subfigure}
    \caption{Finetuned small specialists outperform strong generalists on ReMedE.}
    \label{fig:leaderboard}
\end{figure}

\vspace{-30pt}
\paragraph{Large generalist models underperform on real-world clinical predictive tasks.}
We evaluate both large generalist foundation models and medQA leaderboard models under zero-shot inference, and find that they underperform on \textsc{ReMedE} tasks. After finetuning, \textsc{Lang1-1B} outperform both two finetuned models (\textsc{Llama 3.2 1B} and LoRA \textsc{Deepseek R1 Distill Llama 3.2 70B})  by 3.64\% to 6.75\% AUROC. \textsc{Lang1-1B} also outperforms the best zero-shot performance (including models up to \textsc{DeepSeek R1 671B}) by 1.66\% to 23.66\% AUROC. 

\autoref{fig:leaderboard-zero-shot} shows that on specialized tasks such as insurance denial, both leaderboard models (\textsc{Llama 2 7B}---light blue, \textsc{Llama 3.2 1B}---blue, \textsc{MedMobile}---grey) and our own pretrained models (\textsc{Lang1-1B}---purple) underperform in the zero-shot setting. While mortality prediction has up to 94.2\% AUROC, the other four tasks range between 36.6\%-71.7\% AUROC. This shows that even the strongest generalist models falter on these specialized operational tasks. 

\autoref{fig:leaderboard-finetuning} compares the best zero-shot result per task against a few finetuned models, including \textsc{Lang1} (100M, 1B and 7B), \textsc{Llama 3.2 1B}, and a parameter-efficient finetuned version of \textsc{Deepseek R1 Distill Llama 70B}. Across all five tasks, \textsc{Lang1-1B} and \textsc{Lang1-7B} (purple bars) achieve higher AUROC than the best zero-shot model (magenta) and the other finetuned models (light blue and dark blue). The improvements over the other finetuned models range from 3.64\%--6.75\%. The improvements over the best zero-shot baseline range from 1.66\% (mortality) to 23.66\% (insurance denial), with the largest gains observed at insurance denial prediction, which is \textsc{Deepseek R1}'s  worst performing task. 

\begin{figure}[ht!]
    \centering
    \begin{subfigure}[t]{\textwidth}
    \centering
\includegraphics[width=.6\linewidth]{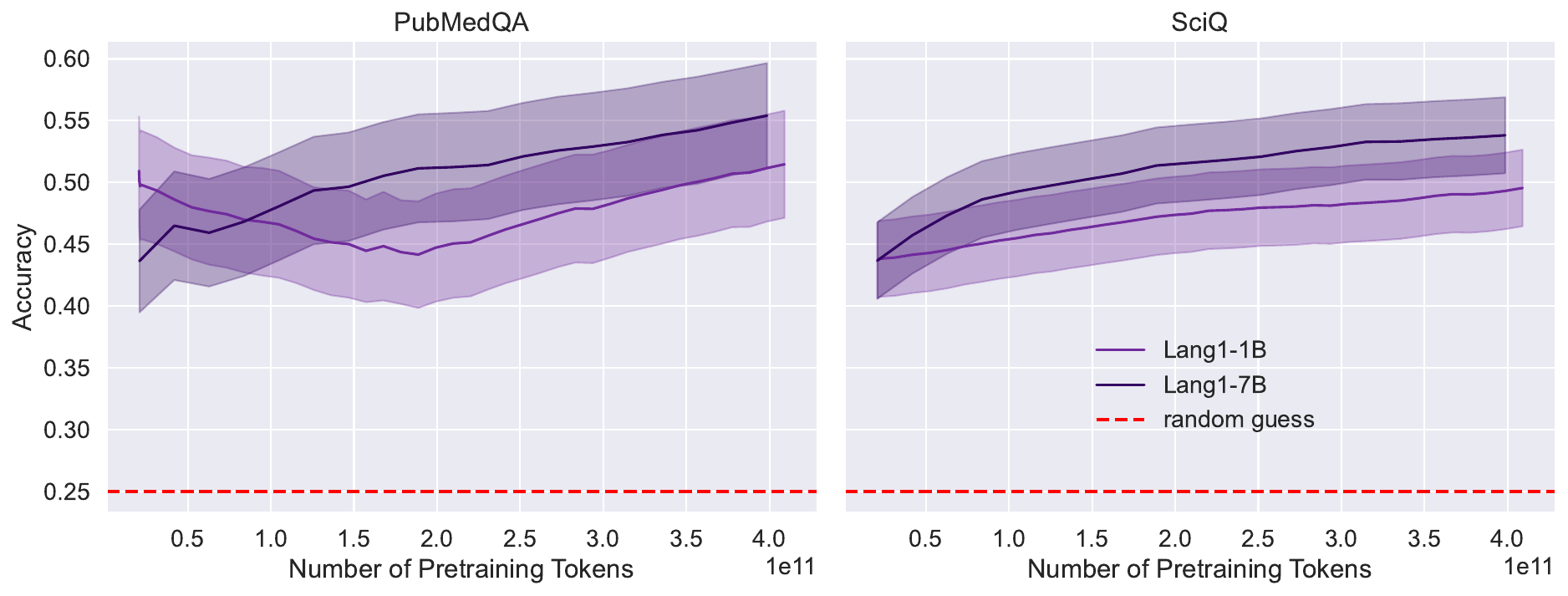}
\caption{Reading comprehension performance increases from pretraining. }\label{fig:emergence_reading}

    \end{subfigure}\hfill
    \begin{subfigure}[t]{\textwidth}
    \centering
    \includegraphics[width=.7\linewidth]{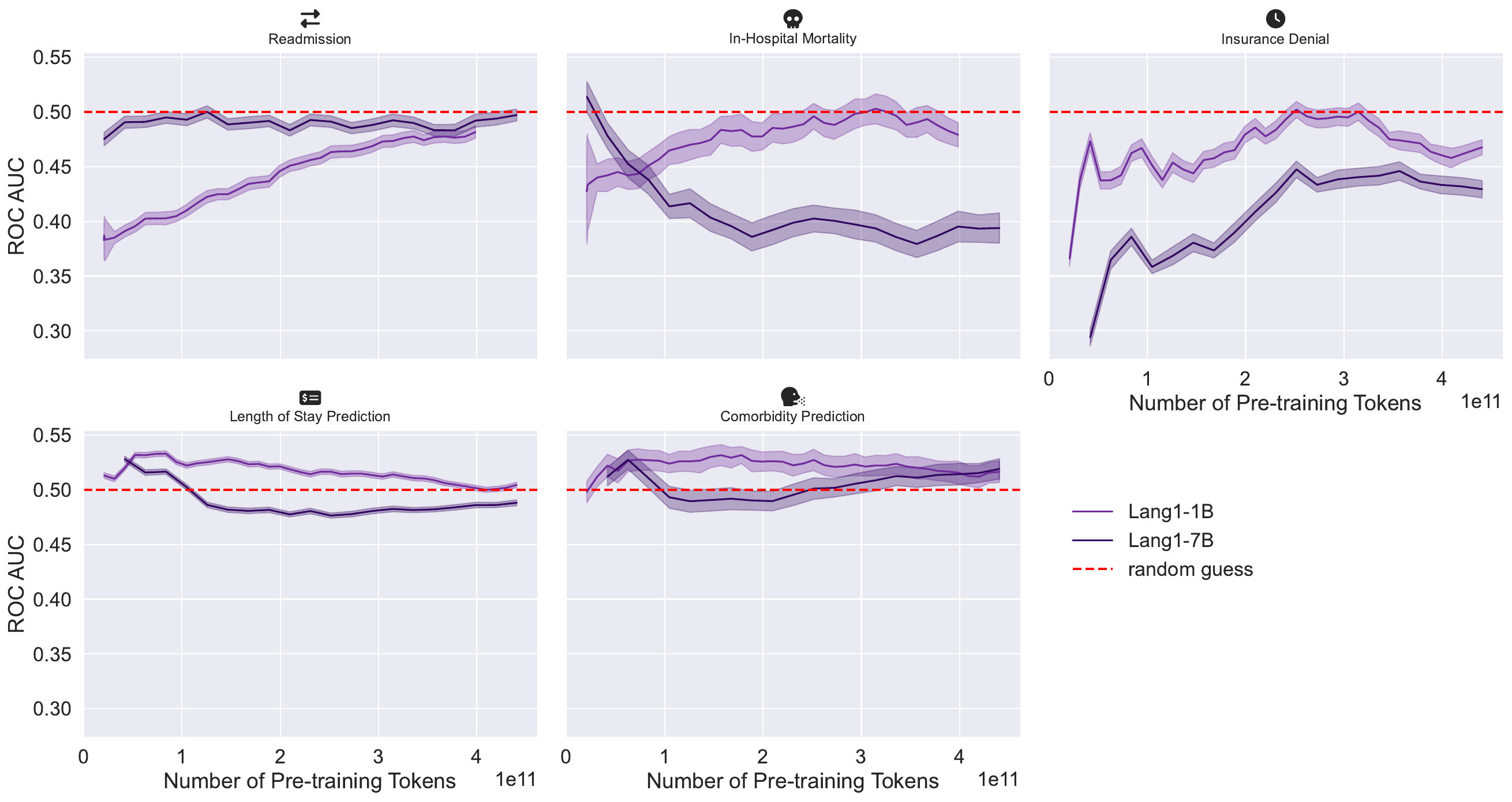}
        \caption{Clinical classification performance does not rapidly emerge from pretraining.}\label{fig:emergence_clinical}
    \end{subfigure}
    \caption{Zero-shot clinical classification performance does not increase over the course of pretraining, unlike reading comprehension.  Error bars depict the 95\% confidence interval.} 
    \label{fig:emergence_pretrain}
\end{figure}

\vspace{-30pt}

\paragraph{Clinical performance does not emerge during pretraining, unlike reading comprehension}
We tracked zero-shot performance of \textsc{Lang1} (1B and 7B) throughout pretraining, as a function of the number of tokens seen. On comprehension tasks (Method~\ref{sec:comprehension_data} for data details), accuracy increased with additional pretraining data (\autoref{fig:emergence_reading}), consistent with the intuition that language models improve on text-based reasoning tasks as they are exposed to more data. In contrast, zero-shot AUROC on clinical classification tasks (\textsc{ReMedE}) remained close to or below random chance across the entire pretraining trajectory (\autoref{fig:emergence_clinical}). We hypothesize that the mapping from clinical notes to outcomes does not emerge from learning next token prediction on unlabeled texts alone, but must be learned through either task-specific finetuning or alternative pretraining objectives. 

\begin{figure}[ht!]
    \centering
    \begin{subfigure}[t]{\textwidth}
        \centering
        \includegraphics[width=.8\linewidth]{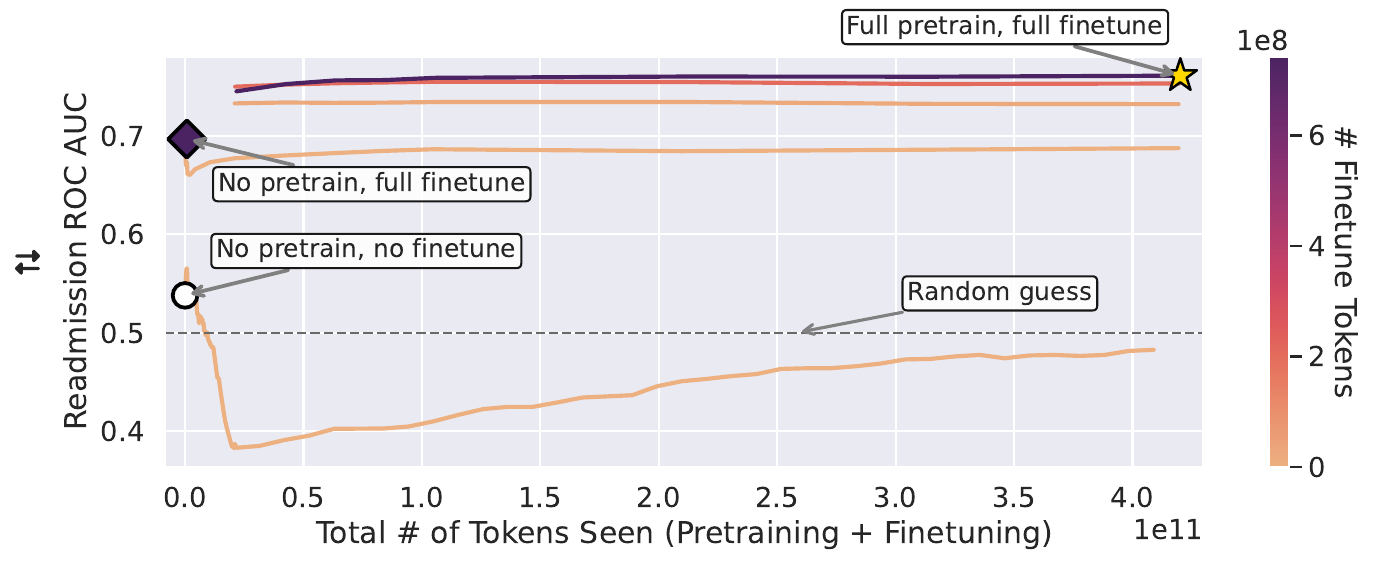}
        \caption{Finetuning is more token-efficient for performance gains, but pretraining still provides value. At any fixed token budget (a vertical slice on the \textit{x} axis), using more finetuning tokens (darker colors) yields a higher ROC AUC for \textsc{Lang1-1B}'s checkpoint trajectory. Nonetheless, a clear gap remains between fully finetuning without pretraining (purple diamond) and the fully pretrained model (yellow star).}
        \label{fig:finetuning-vs-pretraining-efficiency}
    \end{subfigure}
        \vspace{1em}
    \begin{subfigure}[t]{.48\textwidth}
        \centering
        \includegraphics[width=\linewidth]{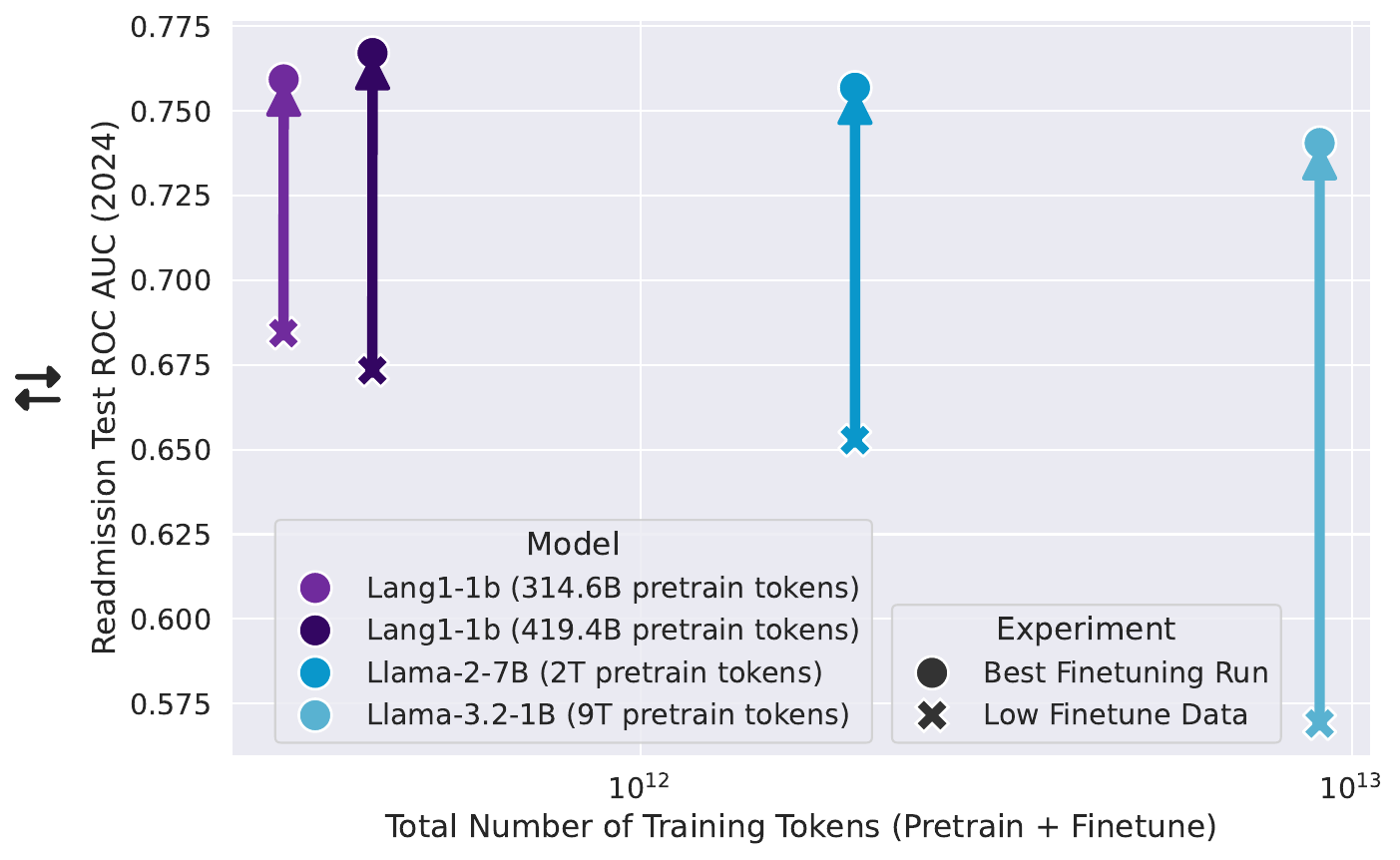}
        \caption{Clinically pretrained models (purple \textsc{Lang1-1b}), when finetuned, outperform generalist models of similar size (blue \textsc{Llama} models), especially in low data regime (cross). Arrows track the same pretrained model as it is finetuned on different numbers of examples.}
        \label{fig:low-vs-full-finetune}
    \end{subfigure}\hfill 
        \begin{subfigure}[t]{.48\textwidth}
        \centering
        \includegraphics[width=\linewidth]{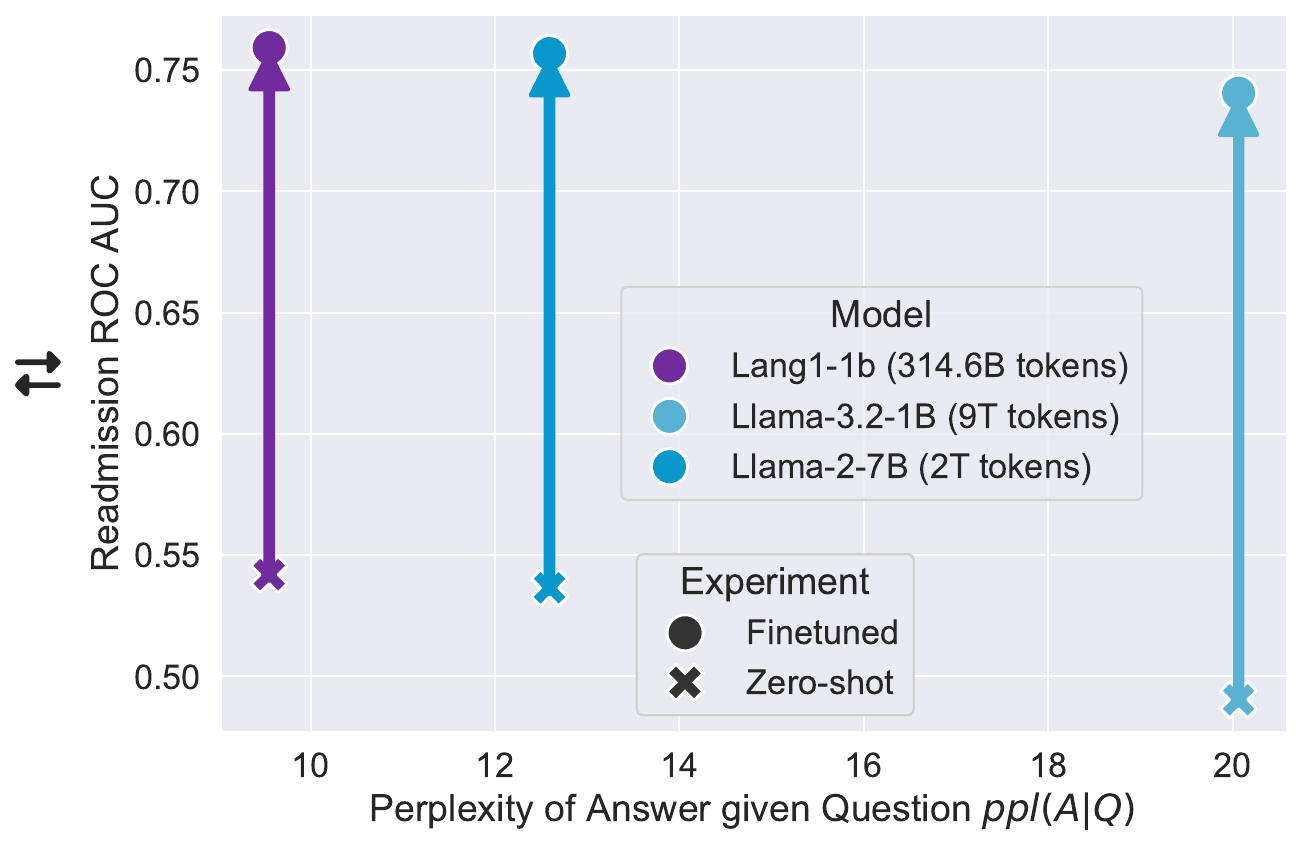}
        \caption{Lower model perplexity on \textsc{ReMedE} answer-question pairs is associated with better zero-shot (cross) and finetuned (circle) performance. \textsc{Lang1}(purple) has lower perplexity despite fewer total pretrain tokens (though more clinical tokens). Arrows track the same model's performance.}
        \label{fig:ppl_zero_shot_finetune}
    \end{subfigure}
    \caption{Finetuning is are more token-efficient than pretraining for performance gains (\autoref{fig:finetuning-vs-pretraining-efficiency}), but in-domain pretraining enables sample-efficient finetuning (\autoref{fig:low-vs-full-finetune}). This advantage is also associated with lower perplexity on in-domain tasks (\autoref{fig:ppl_zero_shot_finetune}).}
    \label{fig:combined-finetuning}
\end{figure}

\vspace{-30pt}

\paragraph{Compute is better spent on finetuning, but pretraining makes finetuning more efficient}
\autoref{fig:finetuning-vs-pretraining-efficiency} examines the pretraining and finetuning trajectory of the \textsc{Lang1-1B} readmission model under a fixed total token budget (pretraining + finetuning, \textit{i.e.}, each vertical slice). During pretraining of \textsc{Lang1-1B}, checkpoints were saved after each one million training tokens. Each pretrain checkpoint is finetuned using 100--362,259 discharge notes with readmission label (2.0M--742.0M tokens). The compute budget is the sum of pretraining token for that checkpoint and the number of finetuning tokens used to finetune that particular checkpoint. Within each slice, increasing the proportion of finetuning tokens (darker colors) consistently improves performance. At the same time, \textbf{pretraining still provides value}: even with maximal finetuning data, models initialized from pretraining (purple diamond) outperform the randomly initialized one (yellow star) by 7.03\% AUROC. A similar pattern appears in \autoref{fig:low-vs-full-finetune}: when finetuning data are scarce, \textsc{Lang1-1B} (purple) outperforms generalist models of comparable scale (blue \textsc{Llama-2-7B} and \textsc{Llama-3.2-1B}) that were pretrained on more nonclinical tokens, demonstrating the efficiency gains of domain-specific pretraining. In fact, \autoref{fig:ppl_zero_shot_finetune} shows that \textsc{Lang1-1B}, despite being trained on fewer tokens, achieves lower perplexity on answer--question pairs and stronger zero-shot and finetuned performance than \textsc{Llama-2-7B} and \textsc{Llama-3.2-1B}. Ablations show that larger models trained on more recent data have better performance (\autoref{sec:pretrain_ablation}).


\begin{figure}[ht!]
    \centering
        \centering
    \begin{subfigure}{0.48\textwidth}
        \includegraphics[width=\linewidth]{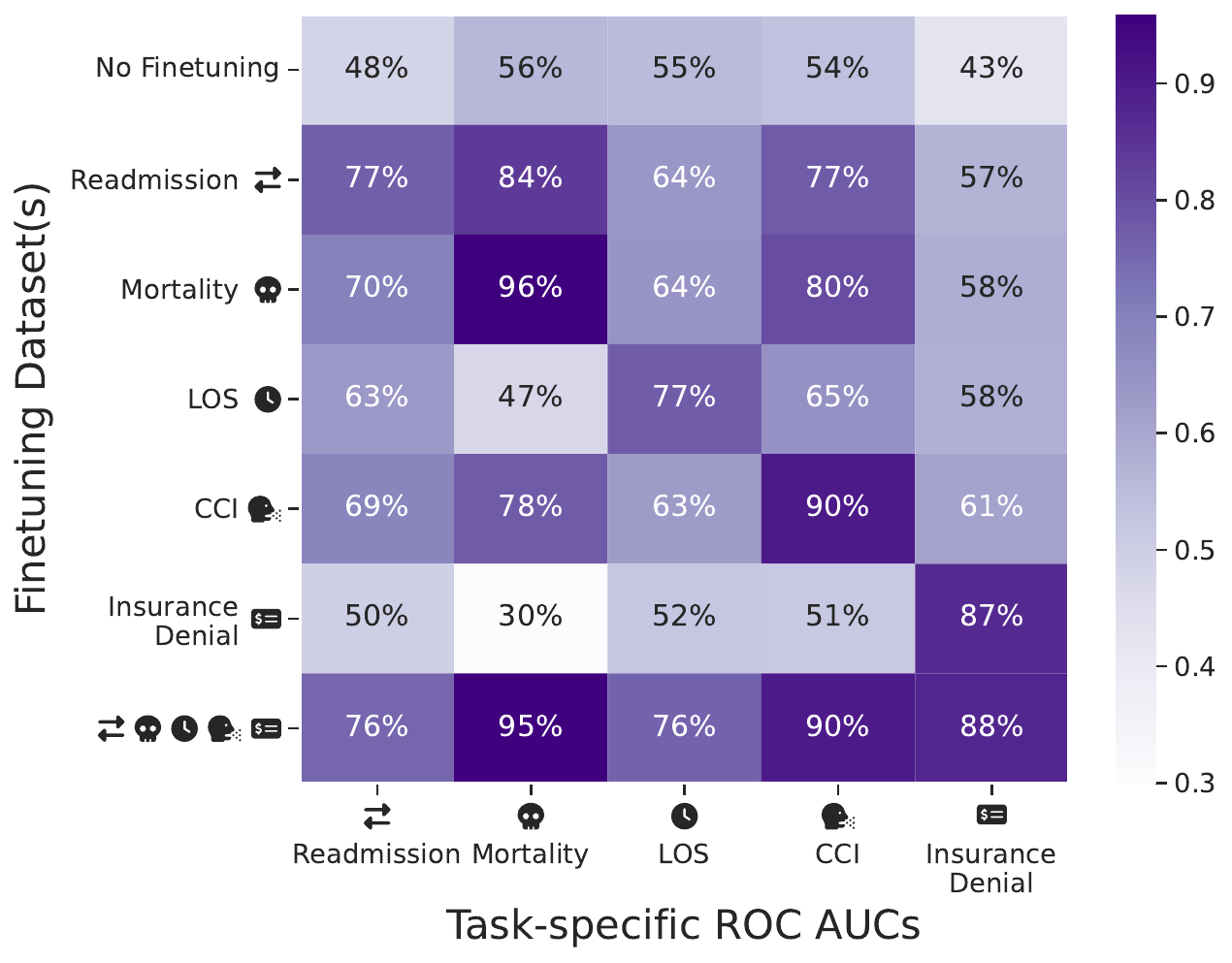}
        \caption{Finetuning \textsc{Lang1-1B} can often transfer across tasks. The heatmap shows the finetuned model's performance when finetuned on a subset of \textsc{ReMedE} tasks ($y$ axis) and evaluated on all five \textsc{ReMedE} tasks ($x$ axis).}\label{fig:heatmap} 
     \end{subfigure}
     \hfill
    \begin{subfigure}{0.48\textwidth}
    \includegraphics[width=\linewidth]{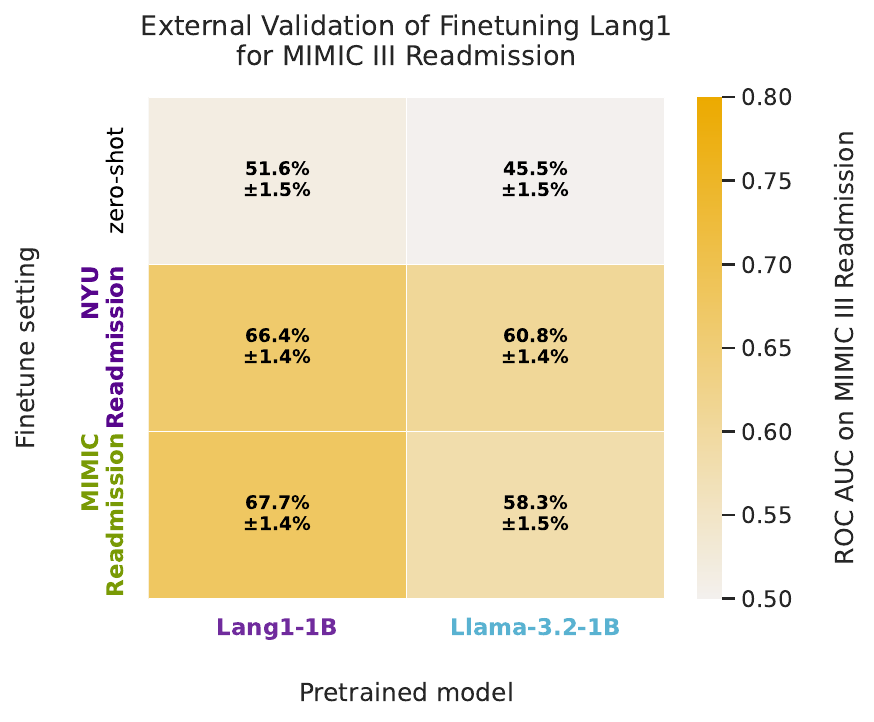}
    \caption{Finetuning on NYU Readmission (purple) transfers well (darker yellow) to MIMIC III Readmission (green). Overall performance is better on finetuning \textsc{Lang1-1B} (purple) compared to \textsc{Llama 3.2 1B} (blue).}\label{fig:mimic_transfer}
    \end{subfigure}
    \caption{\textsc{Lang1} is able to transfer to unseen task (\autoref{fig:heatmap}) and a different health system (\autoref{fig:mimic_transfer}).}
    \label{fig:transfer-learning-heatmap}
\end{figure}

\vspace{-30pt}

\paragraph{\textsc{Lang1} is able to transfer to unseen task and a different health system} The heatmaps in \autoref{fig:heatmap} show \textsc{Lang1-1B} finetuned on one or all ReMedE tasks~(rows) and evaluated on all five tasks~(columns). Overall, Lang1-1B achieves strong single-task~(diagonal) and joint-task~(last row) performance, and is well calibrated~(\autoref{sec:cali}). Many tasks transfer. For example, finetuning on readmission (second row) boosts performance on the other four tasks. However, this transfer could be asymmetric, \textit{i.e.}, mortality helps LOS (third row, third column), but LOS does not help mortality (fourth row, second column), which can be explained using domain knowledge (\autoref{app:asymmetry}). Compared to \textsc{Lang1-1B}, \textsc{Llama-3.2-1B} has overall worse performance and a different transfer pattern (\autoref{app:llama_3_2_transfer_vs_lang1}), suggesting that the pattern depends on the pretrained model. 

\autoref{fig:mimic_transfer} shows \textsc{Lang1-1B} and \textsc{Llama-3.2-1B} finetuned using different readmission data (MIMIC III or NYU) and test on MIMIC. Finetuning \textsc{Lang1-1B} shows better performance on both data. For \textsc{Lang1-1B}, finetuning on MIMIC is slightly better than NYU by 1.2\% AUROC. However, for \textsc{Llama-3.2-1B}, finetuning on NYU is surprisingly better than finetuning on MIMIC by 2.5\% AUROC. We suspect this is because NYU$+$ Readmission has more labeled pairs than MIMIC readmission, suggesting that nonclinical models may benefit more from a large number of slightly out-of-distribution examples compared to a smaller number of in-distribution examples. Indeed, downsampling NYU dataset to the same size as MIMIC would lead to similar results for \textsc{Llama-3.2-1B} (\autoref{fig:downsample_nyu_vs_mimic}). \autoref{app:external_val} extends this analysis to MIMIC mortality and LOS with similar observations.

\section{Discussion}

We present what is, to our knowledge, the first comprehensive study of LLMs for hospital and health-system operations. We detail a full-cycle approach, in which we build, benchmark and test the robustness of these models as part of the their operational deployment within the NYU Langone Health System. Our work has implications for the broader debate between generalist and specialist models, LLM generalizability, and pretraining-finetuning dynamics.


\paragraph{Why Operational Tasks Matter}  
Much of the current excitement in medical AI centers on diagnostic reasoning~\citep{panagoulias_2023_evaluatingchatgpt,goh_2024_diagnostic,Nori2025-va,Tu2025-el,Johri_Jeong_Tran_Schlessinger_Wongvibulsin_Barnes_Zhou_Cai_Van_Allen_Kim_et_al._2025,Arora2025-us}. These are valuable directions, but they do not fully capture the day-to-day challenges physicians face. Healthcare is as much operational as it is clinical. Physicians spend only 26\% of their time in direct patient care, with much of the remainder devoted to documentation, insurance, and scheduling \citep{Sinsky2016-yo}. Operational outcomes such as readmission, insurance denial, and length of stay directly shape costs, capacity, and continuity of care. Predicting them is actionable, and improving them is measurable. By advancing operational tasks alongside diagnostic ones, we can reduce costs, improve care delivery, and make healthcare more accessible. Notably, we find that many of these tasks require specialized finetuning, and are not out-of-the-box accessible to generalist models, likely reflecting their poor representation in web-scale datasets.

\paragraph{On The Need for Real-World Evaluation}  
Our results show the limitations of relying on proxy benchmarks as evidence of readiness for clinical deployment. Several MedQA leaderboard models under-perform on ReMedE, suggesting that proxy benchmark success does not establish clinical utility. Evaluating models directly on real-world, task-specific outcomes is therefore essential, not only to identify models that genuinely improve hospital operations, but also to avoid overestimating the safety or value of systems based solely on proxy measures~\citep{he_2024_lab_results,Hager2024-cz,Yan2025LLMSE}.

\paragraph{The Costs of Training}  
A common concern about specialized models is whether they can be trained affordably outside of industrial labs. Our experience with \textsc{Lang1} suggests that the answer is yes. Training the 1B-parameter model on 314.5B tokens (150k steps) required roughly 30 days on 64 H100s, which at AWS \texttt{p5.48xlarge} pricing corresponds to a cost of about \$180{,}000. While non insignificant, this figure is orders of magnitude lower than the multimillion-dollar budgets required for frontier generalist models \citep{hoffmann2022chinchilla,Luccioni2023-lq, Cottier2024-iv, DeepSeek-AI_Guo_Yang_Zhang_Song_Zhang_Xu_Zhu_Ma_Wang_et_al._2025}, yet delivers substantial performance gains on real-world hospital tasks. For a large health system, such costs are comparable to routine IT infrastructure upgrades and therefore financially and operationally feasible and profitable. This supports recent position papers \citep{Feng2025-kn, Belcak2025-ux} that argue small, specialized models are the future of agentic AI because they can be more reliable, economical, and aligned with domain-specific needs. 

\paragraph{In-House Models}  
From an operational standpoint, training and deploying models in-house can make more sense than relying exclusively on generalist models delivered via commercial APIs. In this paradigm, hospitals not only reduce long-term costs and safeguard patient data, but also retain the ability to adapt models as documentation practices, coding standards, and patient populations shift over time, a need highlighted by our temporal robustness results. In contrast, the dependence on external APIs can introduce ongoing expenses, privacy risks, and limited control over model behavior. Our findings suggest that even modestly sized models, trained locally, can outperform larger off-the-shelf alternatives, reframing clinical AI from ``renting intelligence" to ``building institutional assets" that evolve with the health system itself.

\paragraph{Pretraining Is Not All You Need}  
While the reading comprehension capabilities emerge directly from large-scale pretraining, our finding provides evidence that high-stake objective predictions represent a different class of problem. Strong performance on \textsc{ReMedE} tasks require explicit finetuning and does not emerge from pretraining alone, even with domain-specific data. This reliance on finetuning is shared by general-purpose chatbots, but the underlying tasks are fundamentally different. Chatbot finetuning aligns emergent generalist skills to \textit{subjective}, preference-based goals \citep{Ouyang2022-as}. In contrast, \textsc{ReMedE} tasks require finetuning to build a new, non-emergent predictive skill against an \textit{objective}, grouth-truth target.  We hypothesize that this is because clinical predictions rely on complex relationships not well captured by a next-token-prediction objective. Our experiments show that these tasks demand domain-specific supervision built on top of in-domain pretraining.


\paragraph{Why Transfer Matters}  
Healthcare tasks often suffer from limited labels due to the expertise required for annotation, and in some cases it is practically impossible to obtain large labeled datasets (\textit{e.g.}, for rare conditions). Our experiments show that instruction finetuning enables transfer across related tasks, where supervision on one outcome (\textit{e.g.,} readmission) improves performance on others (\textit{e.g.}, mortality, length of stay). We also showed that finetuning \textsc{Lang1} on NYU can transfer to a different hospital. This capability is especially valuable in clinical settings because it reduces dependence on costly annotation pipelines and makes it possible to tackle tasks where labels are scarce or unattainable, effectively maximizing the value of individual annotations. 

\section{Conclusions}

We introduced \textsc{Lang1}, a family of domain-specialized models trained on EHR and web text. To evaluate \textsc{Lang1}, we developed \textsc{ReMedE}, an operations-grounded benchmark for evaluating language models across five operationally significant tasks. We find that large generalist models underperform \textsc{Lang1-1B} by 1.66\% - 23.66\% AUROC, despite their strong performance on general and proxy medical benchmarks. \textsc{Lang1-1B} also outperformed other finetuned models up to \unboldmath{$70\times$} larger by 3.64\%--6.75\%. These results demonstrate that success on proxy benchmarks does not guarantee effectiveness in deployment-critical settings.

Our analysis shows that clinical prediction capabilities do not arise from pretraining alone; explicit finetuning is necessary, though in-domain pretraining improves data efficiency. \textsc{Lang1} also demonstrates promising transfer across related tasks and health systems, suggesting that carefully designed instruction finetuning can reduce dependency on expensive annotation. Training costs for \textsc{Lang1} are substantially smaller than frontier models and affordable for large health systems, making specialized models both effective and practical.

Future works include expanding \textsc{ReMedE} to additional institutions, more diverse patient populations, and task types beyond classification outcomes. This is important for establishing shared standards for evaluating models in real operational contexts.

Our findings challenge the assumption that ever-larger internet-trained models will generalize to all domains and instead point to a more hopeful direction for clinical AI. In healthcare, where patient safety is paramount, progress must be deliberate and grounded in real operational outcomes. Benchmarks like \textsc{ReMedE} and specialized models such as \textsc{Lang1} show that smaller, domain-specific systems can be accurate, affordable, and adaptable, offering a scalable path forward that directly supports hospital operations and improves patient care. 

\pagebreak
\section{Methods}\label{sec:method}

\subsection{Data Collection}

Data are extracted via structured query language (SQL) scripts to query the NYU Langone EHR, prototyped in an interactive web-based editor (Cloudera Hue), and exported as comma-separated files (CSVs) to an on-prem high-performance computing (HPC) cluster. 

\paragraph{Preprocessing} 
Raw CSV notes (including pathology, radiology, and general hospital notes) are loaded with standard ASCII-encoding using Python Dask \citep{Rocklin2015-se} for distributed processing. We concatenate narrative fields, standardize punctuation, spacing and formatting via regular expression substitutions, remove non-ASCII and malformed characters, remove errant whitespace and newlines, and filter out short notes (less than 10 words or with placeholder values such as $<$NA$>$). 

\subsection{Datasets}

\subsubsection{Pretraining Dataset}\label{sec:pretrain_data}

\paragraph{Web texts} We use SlimPajama (627B tokens) \cite{cerebras_2023_slimpajama}, a large extensively deduplicated, multi-corpora, open-source dataset for training LLMs. Its sources include Commoncrawl \citep{commoncrawl2024}, C4 \citep{c4dataset2020}, Github, Books \citep{bookcorpus2,rae2020pg19}, Arxiv, Wikipedia and StackExchange.

\paragraph{NYU Notes}
This dataset is previously created using unlabeled inpatient hospital notes signed by medical professionals from the NYU Langone Health EHR\footnote{This study is approved by the Institutional Review Board (IRB) at NYU Langone Health. The methods are carried out in accordance with the IRB's relevant guidelines and regulations.} for patient encounters starting from January 2011 to May 2020. NYU Notes contains 387,144 patients, 7,247,694 notes, and 4,112,249,482 words in total. NYU Notes are used to train and evaluate NYUTron. \cite{Jiang2023-ig}

\paragraph{NYU Notes$+$} 
This dataset builds on NYU Notes by including a wider range of note types and covering a longer time span, resulting in a total word count 14.5 times greater than that of NYU Notes. NYU Notes$+$ contains unlabeled hospital notes, pathology note and radiology notes from NYU Langone Health EHR from 2003 to 2023. It comprises 11,689,342 patients, 180,487,092 notes, and 59,917,646,788 words.

\subsubsection{Finetuning Datasets and ReMedE Test Set}\label{sec:finetune_data}

We derive five task-specific labeled datasets by combining NYUTron \cite{Jiang2023-ig} finetune datasets, with the addition of 2024 temporal test set to approximate deployment robustness. The 2024 temporal tests set are used for ReMedE. See \autoref{sec:timeline_viz} for a visualization of the data split timeline and \autoref{app:remede_stats} for detailed dataset statistics. \autoref{app:patient_control} shows that a small percentage of patient overlap does not over-estimate model performance on readmission. For both zero-shot evaluation and finetuning (section \ref{sec:ft}), the dataset were converted to multiple choice format~(\autoref{sec:detailed_prompts}).

\paragraph{NYU$+$ Readmission} Readmission occurs when a patient returns to the hospital shortly after discharge.  Predicting readmissions is critical for identifying patients who need longer stay or post-discharge support, and it also serves as a key hospital quality metric. This finetuning dataset contains discharge notes with 30-day all-cause readmission labels. The notes comprise of a subset of NYU$+$ Notes whose encounters end between January 2013 and November 2021, and additional discharge notes from 2024 for the temporal test. Rehabilitation, dialysis and palliative care notes are excluded to focus on modeling acute readmission. A positive label is assigned if the patient is readmitted within 30 days of discharge, and a negative label otherwise. We split the dataset into five sets. The first three sets are train, validation, and test set with a 8:1:1 ratio from 2013 to May 2021. The 2021 temporal test includes notes from June to December 2021. The 2024 temporal test set includes notes from 2024. The positive class ratio range from  10.81\% to 11.29\% across these five sets. The dataset contains 421,429 patients, 604,326 notes and 607,877,177 words in total.

\paragraph{NYU$+$ In-Hospital Mortality} In-hospital mortality prediction identifies patients at highest risk of death during their admission, enabling timely palliative care consultations and goals-of-care discussions that align treatment with patient prognosis. This finetuning dataset contains History and physical (H\&P) notes with in-hospital mortality labels. The notes comprise of a subset of NYU$+$ Notes for encounters ending between January 2013 and November 2021, with additional H\&P notes from 2024 for the temporal test. A positive label is assigned is the discharge disposition is ``Expired", and a negative label otherwise. We split the dataset into five sets. The first three sets are train, validation, and test set with a 8:1:1 ratio from 2013 to May 2021. The 2021 temporal test includes notes from June to December 2021. The 2024 temporal test set includes notes from 2024. The positive class ratio range from 1.78\% to 1.93\% across these five sets. In total, the dataset contains 395,991 patients, 566,748 notes, and 608,603,182 words. 

\paragraph{NYU$+$ Length of Stay (LOS)} LOS is the number of days a patient remains hospitalized. Predicting LOS is essential for bed management, staffing allocation, and discharge planning. This finetuning dataset contains H\&P notes with label for binned length of stay. The dataset comprises of a subset of NYU$+$ Notes for encounters ending between January 2013 and November 2021, with additional H\&P notes from 2024 for the temporal test. We assign the labels based on quantile. We assigned label 0 for an LOS below the 25\% quantile (0 to 2 days), label 1 for an LOS between the 25\% and 50\% quantile (3 days), label 2 for an LOS between the 50\% and 75\% quantile (4 to 5 days) and label 3 for an LOS above the 75\% quantile ($>5$ days). We split the dataset into five sets. The first three sets were train, validation, and test set with a 8:1:1 ratio from 2013 to May 2021. The 2021 temporal test includes notes from June to December 2021. The 2024 temporal test set includes notes from 2024. The majority class ratio (0 to 2 days) range from  41.49\% to 45.64\% across these five sets. The minority class ratio (more than 5 days) range from 23.92\% to 26.34\%. In total, the dataset contains 395,991 patients, 566,748 notes and 608,603,182 words. 

\paragraph{NYU$+$ Insurance Denial} Insurance denials occur when payers reject claims for hospital services. Predicting denials allows hospitals to proactively address documentation gaps, reducing administrative burden and preventing unexpected out-of-pocket costs for patients. This finetuning dataset contains H\&P notes with insurance denial label. The notes comprise of a subset of NYU$+$ Notes whose encounters ends between May 1, 2021 and April 30, 2022, with additional H\&P notes from Janurary 2024 for the temporal test. A positive label is assigned if claim status is ``final, adverse determination" (initial rejection by insurance and again rejected following appeal) or ``final, favorable determination" (initial rejection by insurance and approved following appeal), an negative label otherwise. We split the dataset into five sets. The first three sets are train, validation, and test set with a 8:1:1 ratio from May 2021-Feb 2022. The 2022 temporal test include notes from March-Apr 2022. The 2024 temporal test set includes notes from January 2024. The positive class ratio range from 12.01\% to 13.90\%. The dataset contains 87,974 patients, 97,837 notes, and 89,147,715 words.

\paragraph{NYU$+$ Charlson Comorbidity Index (CCI)} 
CCI is a standard score used to quantify a patient's chronic illness based on their medical history \citep{charlsonNewMethodClassifying1987}. It is useful for supporting accurate risk stratification for patients with limited historical data. However, this heuristic score cannot be computed when a patient's medical history is unknown. This dataset addresses this problem by providing History \& Physical (H\&P) notes paired with their corresponding binned CCI scores. It allows models to be trained to impute the CCI score directly from unstructured clinical text. The target CCI is calculated using ICD code associated with the encounter in the EHR following the scoring function in \citep{cci-md-calc}. Encounters with no associated ICD codes are excluded. The CCI is discretized into five classes: comorbidity index of 0 ($<$50th percentile), comorbidity index of $1-2$ ($50-75$\% percentile), comorbidity index of $3-4$ (75-90th percentile), and comorbidity index of $5-7$ (90-99th percentile), and comorbidity index $>$7 ($>$99th percentile). We split the dataset into five sets. The first three sets are train, validation, and test set with a 8:1:1 ratio from 2013 to May 2021. The 2021 temporal test includes notes from June to December 2021. The 2024 temporal test set includes notes from 2024. The majority class (score 0) ratio range from 68.40\% to 69.47\%. The minority class (score more than 7) ratio range from 0.059\% to 0.17\%. In total, the dataset has 306,741 patients, 443,915 notes, and 524,739,038 words. 

\subsubsection{External Validation Datasets}\label{app:mimic_datasets}

We also create external validation datasets from MIMIC III \citep{Johnson2023-zf}, which is sourced from Beth Israel hospital in Boston Massachusetts.

\paragraph{MIMIC III Readmission} The labeled dataset has 6\% positive labels, with 52,725 examples and a 70\% train, 15\% validation and 15\% test split. More dataset construction details are in \cite{Yang2022-ws}'s Appendix A. 

\paragraph{MIMIC III Mortality} The labeled dataset has 10.55\% positive labels, with 5658 examples and 80\% train, 10\% validation and 10\% test split.  The dataset is constructed from MIMIC-III clinical notes. As no explicit "Admission Note" label exists, we identify notes by filtering descriptions for "admission" while excluding "readmission" ($\approx$19K notes). To prevent patient bias, we select a single note per hospital stay using a prioritization heuristic. The heuristic first prioritizes Physician Note > General Note > Nurse Note, then prioritizes "resident/attending" > "resident" > "attending" > "fellow" > "h\&p". Finally the heuristic prefers longer notes. This filters down to $\approx$6K notes. We further refine the dataset by removing notes written $>120$ hours after admission or associated with a negative length of stay. The final dataset contains 5,658 unique admission notes. 

\paragraph{MIMIC III LOS} The labeled dataset uses the same 5,658 admission notes as the mortality task, with a mean LOS of 7.96 days. Similarly, the split is 80\% train, 10\% validation and 10\% test. The continuous LOS values are discretized into bins. We adapt the NYU$+$ LOS scheme (Methods \ref{sec:finetune_data}) to handle MIMIC-III's continuous range by treating the original integer bins as upper bounds. For example, the "3 days" bin is modified to capture the continuous range $2 < \text{LOS} \leq 3$.

\subsubsection{Comprehension Datasets}\label{sec:comprehension_data}

We evaluate the performance of \textsc{Lang1} checkpoints on comprehension datasets to analyze the emergence of its nonclinical abilities. 

\paragraph{SciQ \citep{Welbl2017-uh}} The dataset contains 13.7K multiple choice science exam questions with contexts. An example question is \texttt{Mesophiles grow best in moderate temperature, typically between 25\(^\circ\)C and 40\(^\circ\)C (77\(^\circ\)F and 104\(^\circ\)F). Mesophiles are often found living in or on the bodies of humans or other animals. The optimal growth temperature of many pathogenic mesophiles is 37\(^\circ\)C (98\(^\circ\)F), the normal human body temperature. Mesophilic organisms have important uses in food preparation, including cheese, yogurt, beer and wine. \textbackslash n Question: What type of organism is commonly used in preparation of foods such as cheese and yogurt? \textbackslash n Answer:}

\paragraph{PubmedQA \citep{Jin2019-oy}} The dataset contains 1k expert-annotated biomedical question answering dataset collected from PubMed abstracts. An example question is \texttt{Abstract: {Complex regional pain syndrome type I is treated symptomatically ... Early cast-related complaints predicted the development of complex regional pain syndrome (relative risk, 5.35; 95\% confidence interval, 2.13 to 13.42)}\textbackslash n Question: {Can vitamin C prevent complex regional pain syndrome in patients with wrist fractures?}\textbackslash n Answer:}

\subsection{Pretraining \textsc{Lang1}}

We pretrain a family of Llama-style decoders (\textsc{Lang1-100M}, \textsc{Lang1-1B}, \textsc{Lang1-7B}) on a mixture of web texts and NYU Notes+ (section~\ref{sec:pretrain_data}) using next token prediction (\autoref{fig:overview}~a,b).   Detailed demographic statisitics is in Appendix~\ref{sec:nyunotes+_stats}. Unless otherwise noted, \textsc{Lang1} models are trained with equal sampling from both clinical and general sources, which is supported by our pretraining ablations (\autoref{sec:pretrain_ablation}). For tokenization, we use the \textsc{Llama-2-7B} tokenizer (SentencePiece, 32k vocabulary). The 100M-parameter model follows the \textsc{Smol-Llama-101M} architecture with a 1024 context length; the 1B model follows \textsc{TinyLlama-1.1B} with a 2048 context length; and the 7B model follows \textsc{Llama-2-7B} with a 4096 context length

We pretrain on 8 to 64 nVidia 80GB H100s with NVLink. We used the LitGPT \citep{litgpt-2023} library and Fully Sharded Data Parallel \citep{Zhao2023-tm}.

We run a few trials of manual hyperparameter search based on speed, performance and training stability. For all models we use AdamW optimizer with linear warmup (2000 steps), beta1 of 0.9, beta2 of 0.95, eps of 1e-8 and cosine cycle decay down to a minimum learning rate of 4e-5. We use a seed of 3407, weight decay of 0.1, and gradient clipping of 1. In terms of FSDP sharding, we shard gradient and optimizer for models up to 1B, and full sharding for 7B model. For effective batch size, we use 4096 for 100M model for controlled comparison with NYUTron \citep{Jiang2023-ig}, and 1024 for 1B and 7B models.

We implement a monitoring pipeline that automatically triggered few-shot evaluations and generations at fixed intervals of pretraining steps. Slack Weights \& Biases (W\&B) alerts are configured to report loss spikes. Upon detection of anomalies in loss or output, we revert to the most recent stable checkpoint. Validation loss is computed periodically on the held-out 0.1\% validation split and used to determine early stopping and model checkpoint selection.

\subsubsection{Pretrained Models}

We pretrain the following variants of \textsc{Lang1} models (\autoref{tab:model_specs}). Ablations (\autoref{sec:pretrain_ablation}) show that larger models trained on more clinical data has better performance, and mixing in web texts does not hurt much. When we refer to \textsc{Lang1} without specifying data sources, we mean the one trained with NYUNotes$+$ and WebTexts. For instance, \textsc{Lang1-1B} refers to \textsc{lang1-1B-NYUNotes+,WebTexts}.

\begin{table}[ht]
\centering
\caption{Pretrained Model Specifications}
\label{tab:model_specs}
\begin{tabular}{l c l}
\toprule
\textbf{Model Name} & \textbf{Model Size} & \textbf{Pretrain Data} \\
\midrule
\textsc{lang1-100m-NYUNotes} & 100m & NYUNotes \\
\textsc{lang1-100m-NYUNotes+} & 100m & NYUNotes+ \\
\textsc{lang1-100m-NYUNotes+,WebTexts} & 100m & NYUNotes+,WebTexts \\
\textsc{lang1-1B-NYUNotes} & 1B & NYUNotes \\
\textsc{lang1-1B-NYUNotes+} & 1B & NYUNotes+ \\
\textsc{lang1-1B-NYUNotes+,WebTexts} & 1B & NYUNotes+,WebTexts \\
\textsc{Lang1-7B-NYUNotes+,WebTexts} & 7B & NYUNotes+,WebTexts \\
\bottomrule
\end{tabular}
\vspace{-10pt}
\end{table}

\subsection{Finetuning}\label{sec:ft}

We finetune \textsc{Lang1} models (and their trajectories of checkpoints) and other pretrained models (\autoref{tab:other_models}) on ReMedE tasks using multiple choice format (\autoref{fig:overview} panel c). The labeled clinical notes are converted to multiple choice format (\autoref{sec:detailed_prompts}), and we train the model to predict the correct multiple choice option. For fair comparison with NYUTron, we right truncate all clinical notes to a maximum of 512 tokens. All finetuning jobs are done one node of 8 nVidia 80GB H100s with NVLink.

\begin{table}[ht]
\centering
\caption{Additional Model Specifications}\label{tab:other_models}
\label{tab:model_specs_additional}
\begin{tabular}{l c l}
\toprule
\textbf{Model Name} & \textbf{Model Size} & \textbf{Pretrain Data} \\
\midrule
\textsc{llama-3.2-1b} & 1B & Unnamed public mix (9T tokens) \\
\textsc{llama-2-7b} & 7B & Unnamed public mix (2T tokens) \\
\textsc{DeepSeek-R1-Distill-Llama-70B} & 70B & Unnamed public mix (2T tokens) + reasoning data (800k samples) \\
\bottomrule
\end{tabular}
\vspace{-10pt}
\end{table}

Before each full finetuning, we run 5 hyperparameter search trials up to 100 steps using Hydra and Optuna. We random search learning rate (based on \cite{Bengio2012-ok}'s recommendation) in log scale over the closed interval 1e-6 to 1e-3. We used an AdamW optimizer with beta1 of 0.9, beta2 of 0.999, eps of 1e-5. We used a weight decay of 0.02 and no gradient clipping. We used a cosine annealing scheduler with no warmup and a max steps of 5120. The best trial is selected based on both maximum validation AUROC and minimum validation loss. 

For full finetuning, we used the best learning rate from hyperparameter search, and train for maximum 5120 steps with early stopping based on Micro-AUROC and a patience of 300 steps. The probabilities for calculating AUROC is obtained via normalizing the logits of the multiple choice options (e.g., ``A''). We train all parameters except for \textsc{DeepSeek-R1-Distill-Llama-70B}, which has to be finetuned using low-rank adaptation finetuning \citep{Hu2021-fh}  to meet the memory constraint (\autoref{app:lora}). 

For multitask finetuning, we mix examples from each task evenly within each training batch. We increase the total training steps scaled by the number of tasks.

\subsection{Evaluation}

\paragraph{Pretraining evaluation} We monitored the token-level cross entropy loss and perplexity for both training and validation.

\paragraph{Zero-shot and few-shot evaluation} ReMedE is built on LM Eval Harness \citep{Biderman2024-uq}. We implemented the tasks as multiple choice questions and AUROC score as a metric. We implemented a child class of \texttt{LocalCompletionsAPI} to connect on-prem models. For models whose logits are not accessible (e.g., on-prem GPT-4o), we implemented custom sampling function to approximate (\autoref{sec:prob_approx}) the probability via counting choices from 10 generations using a temperature of 1.

\paragraph{Finetuning evaluation} We collected the logits of the multiple choice options, normalized it as probabilities, and calculated AUROC  using sklearn's implementation (same as ReMedE's AUROC backend for consistency). For multiclass classification, we used  One-Versus-Rest (OVR) AUROC.

\paragraph{Uncertainty} To capture the uncertainty arose from the randomness of our test set, we calculated 95\% confidence intervals (CI = $\pm$ 1.96 x SD) by resampling each test set 1000 times using the quantile bootstrap method from scipy. 

\paragraph{Temporal Shift} To better mimic deployment conditions under temporal distribution shift, all AUROCs are reported on test data from 2024 -- drawn from a period after the pretraining data -- unless otherwise noted.  See Appendix \ref{sec:timeline_viz} for a visualization of the test timeline. 

\paragraph{Generalist Models}
We compared against generalist frontier models, including \textsc{DeepSeek R1} (served via vLLM \citep{Kwon2023-jk}), \textsc{DeepSeek R1 Distilled Llama~70B} (vLLM), \textsc{Llama~3.3~70B~Chat} (vLLM), and on-premises \textsc{GPT-4o} (Azure-hosted service). We also evaluated MedQA leaderboard models using in-memory inference, such as \textsc{Llama~3.2~1B}, \textsc{Llama~2~7B}, and \textsc{MedMobile}.

\backmatter

\bmhead{Acknowledgements} E.K.O. is supported by the National Cancer Institute's Early Surgeon Scientist Program (3P30CA016087-41S1) and the W.M. Keck Foundation. L.Y.J. is supported by Apple AIML PhD fellowship. L.Y.J. and A.C. are supported by NSF Award 1922658. K.C., E.K.O., L.Y.J. and A.C. are supported by Institute for Information \& communications Technology Promotion (IITP) grant funded by the Korea government (MSIT) (No. RS-2019-II190075 Artificial Intelligence Graduate School Program (KAIST); No. RS-2024-00509279, Global AI Frontier Lab). We would like to acknowledge J. Golfinos, whose vision and support made this project possible. We would like to acknowledge Michael Costantino, Ph.D., Ali Siavosh-Haghighi, Ph.D., Kevin Yie, M.S., Neelima Sharma, Tedum Sampson from the NYU Langone High Performance Computing (HPC) team. Without their tireless assistance in building and maintaining our GPU cluster none of this research would have been possible. We would also like to thank Dr. Dafna Bar-Sagi,Ph.D., and Nader Mherabi whose support for this research has made everything possible. Thanks to He He, Ph.D., Eunsol Choi, Ph.D., Carlos Fernandez-Granda, Ph.D., Julia Kempe, Ph.D., Vasant Dhar, Ph.D., Keunwoo Choi, Ph.D., Jesse Swanson, Gavin Zihao Yang, William Merrill, Ph.D., Nicholas Lourie, Sophie Hao, Ph.D., Vishakh Padmakumar, Ph.D., Michael Hu, Robert J Steele, Yueying Li, Yunzhen Feng, Guillermo Sapiro, Ph.D., Oussama Elaqchar, Kai Xu, Varun Yerram, Itay Itzhak, Jeff Hammerbacher for their valuable discussions. 

\section*{Declarations}

\bmhead{Ethical approval}
This study was approved by the Institutional Review Board (IRB) at NYU Langone Health (study protocol s21-01189). The methods were carried out in accordance with the IRB's relevant guidelines and regulations.

\bmhead{Data Availability}
The clinical data used for the pretraining, finetuning, validation, and test sets were collected from the NYU Langone Health System EHR maintained by the NYULH Datacore team. Text data was stripped of rich text features and directly included in the dataset "as-is", and was augmented with structured features where noted. It consists of the production medical records of NYU Langone and cannot be made publicly available. For the external validation task, the datasets were obtained from MIMIC III, and are publicly available from their  \href{https://physionet.org/content/mimiciii/1.4/}{website}.

\bmhead{Code Availability}
This work uses several open-source libraries including \href{https://pytorch.org/}{PyTorch}, \href{https://github.com/Lightning-AI/litgpt}{LitGPT}, \href{https://huggingface.co/docs/transformers/index}{Transformers library}, \href{https://github.com/EleutherAI/lm-evaluation-harness}{LM Eval Harness}, and \href{https://github.com/facebookresearch/hydra}{hydra}. Our experimental framework involves the utilization of these libraries and in some cases modification of them. We will release code to replicate the pretraining, finetuning, and testing of the models described in this paper at the time of publication. We include detailed methods and implementation steps in the Methods and Supplementary Information to allow for independent replication.
\bmhead{Author's contribution}
E.K.O. and K.C. supervised the project. L.Y.J and X.H. collected pretrain, finetune and evaluation data (except NYU+ Insurance Denial and MIMIC-III). L.Y.J. and A.C. engineered the software for pretrain and finetune. A.C., X.H., L.Y.J. and J.Z. engineered the software for evaluation. L.Y.J., A.C., X.H., R.D., X.C.L. ran experiments. L.Y.J., A.C., X.C.L., X.H., J.Z., R.D. and K.C. debugged and tested the software. A.C., L.Y.J., K.C., X.C.L., R.S., A.A., K.L.S, Y.C., and Q.P. created figures. A.C., L.Y.J., K.C., E.K.O. conceptualized the training dynamics and transfer experiments. J.S. hosted deepseek and llama 70b inference server. K.E. collected NYU+ Insurance Denial data. Q.P. and L.Y.J. check pretrain data quality and cleaned the data. F.W. processed the MIMIC-III Mortality and LOS data. A.C., L.Y.J, R.D., Y.C., D.A. and J.V.L. reviewed related literature. K.C., E.K.O., Y.A. provided guidance and feedback throughout the project. L.Y.J., A.C., X.H., R.D., A.A., D.A., J.V.L., Q.P., Y.C., R.J.S., K.C., E.K.O. wrote the initial draft. All authors edited and revised the manuscript.

\textbf{Conflict of interest}
E.K.O. reports consulting with March AI, Sofinnova Inc., Google Inc., income from Merck \& Co., and Mirati Therapeutics, and equity in Artisight Inc. K.C. is employed by Prescient Design, a subsidiary of Roche. A.C. is employed by Google Deepmind. Q.P. is employed by Faction Imaging Inc. J.S. is employed by March AI. There are no other potential conflicts of interest. The work presented herein was performed exclusively within the NYU Langone Health System.

\newpage

\begin{appendices}
\section{Data Timeline}\label{sec:timeline_viz}

\paragraph{\textsc{Lang1}'s pretrain data covers a wider time window than \textsc{NYUTron}\citep{Jiang2023-ig}} \textsc{NYUTron}'s pretrain data is from 2013 to May 2021. \textsc{Lang1}'s pretrain data covers a longer span, from 2003 to 2023. Overall, \textsc{Lang1}'s pretrain corpus is more than 10 times the size of NYUTron.

\paragraph{\textsc{Lang1} uses the same finetuning dataset for training as \textsc{NYUTron}, but adds additional temporal test} For both \textsc{NYUTron} and \textsc{Lang1} finetuning, we have a temporal test set to better mimic the deployment scenario, where the test set comes from the future of training set. For \textsc{NYUTron}, the temporal test set is between June to December of 2021. For \textsc{Lang1}, the temporal test set is in 2024. All performance we report in main is performance on 2024 temporal test set, unless otherwise specified.

\vspace{-10pt}
\begin{figure}[htbp]
    \centering
    \includegraphics[width=\linewidth]{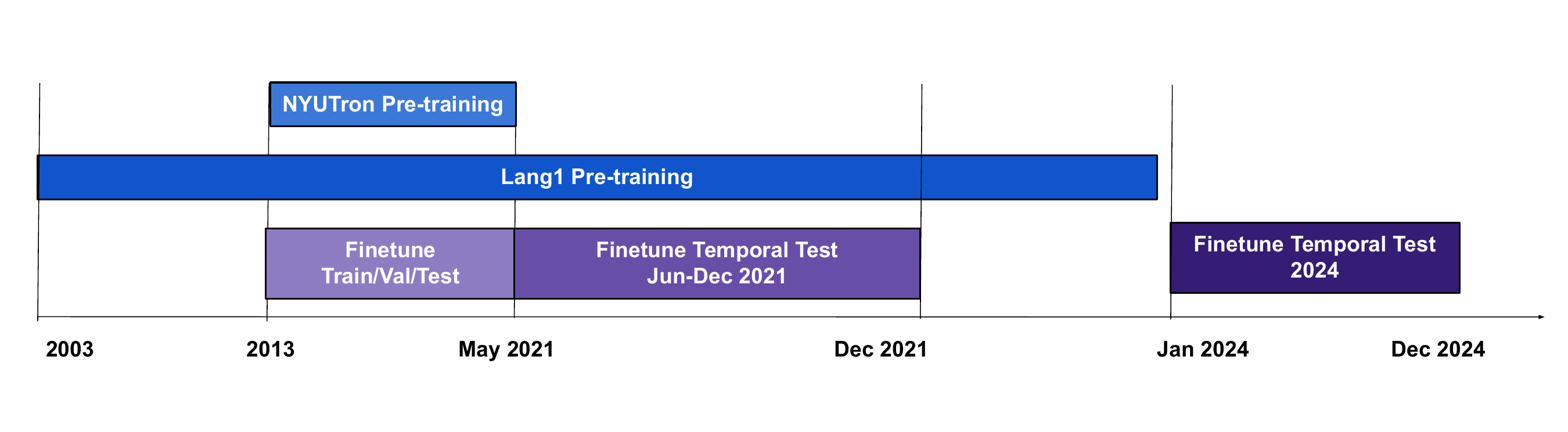}
    \caption{Illustration of timeline of pretrain and finetune dataset for both \textsc{Lang1} and \textsc{NYUTron}. \textsc{Lang1} covers a wider time window for pretraining and added additional temporal test set in 2024 to capture temporal distribution shift.}
    \label{fig:placeholder}
\end{figure}\

\vspace{-30pt}

\begin{figure}[h]
    \centering
    \includegraphics[width=0.5\linewidth]{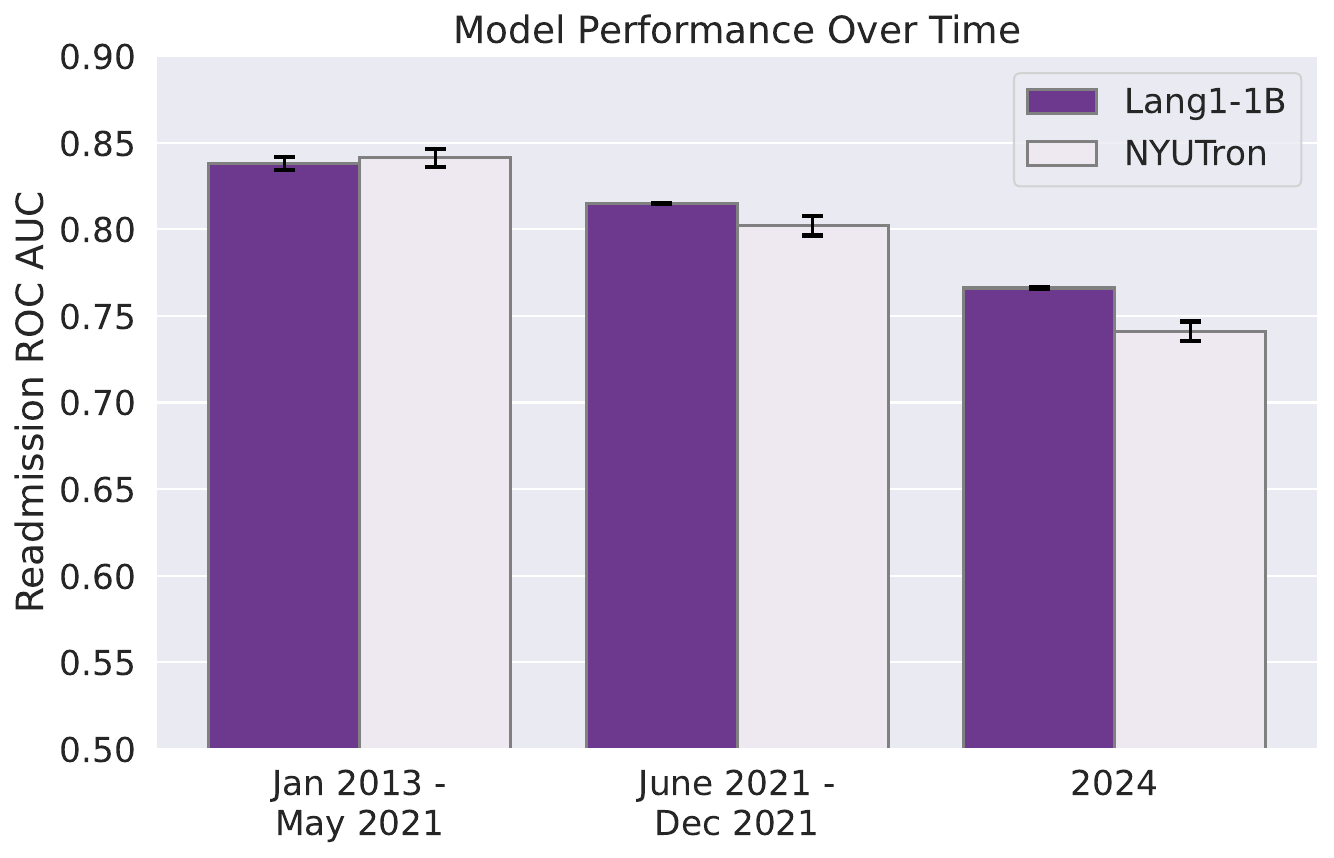}
    \caption{Models perform worse on temporal test and exhibit different level of degradation in face of temporal shift.}
    \label{fig:deterioration}
\end{figure}

\paragraph{Temporal test is important and difficult} \autoref{fig:deterioration} shows that both \textsc{Lang1-1B} (purple) and \textsc{NYUTron} (pink) perform worse as test data are sampled from a further future, illustrating the importance of evaluating with temporal test to capture deployment scenario.

\newpage
\section{ReMedE Dataset Statistics}\label{app:remede_stats}

The following tables show the note counts (\autoref{tab:note_counts}) and patient counts (\autoref{tab:distinct_patients}) across each split, and class ratio for each task: readmission (\autoref{tab:readmission_ratios}), mortality (\autoref{tab:mortality_ratios}), length of stay (\autoref{tab:los_ratios}, comorbidity imputation (\autoref{tab:cci_ratios}, and Insurance denial (\autoref{tab:insurance_ratios}).

\begin{table}[!htbp]
\centering
\caption{Distinct note counts for each ReMedE task across five splits.}
\begin{tabular}{lcccccc}
\toprule
\textbf{Task} & \textbf{Train} & \textbf{Val} & \textbf{Test} & \textbf{Temporal Test 2021} & \textbf{Temporal Test 2024} & \textbf{Total} \\
\midrule
Readmission      & 362,259 & 45,282 & 45,283 & 53,916 & 97,586 & 604,326 \\
CCI              & 256,676 & 32,085 & 32,085 & 42,137 & 80,932 & 443,915 \\
Length of Stay   & 334,515 & 41,814 & 41,815 & 51,018 & 97,586 & 566,748 \\
Insurance Denial & 41,842  & 2,325  & 2,325  & 9,299  & 42,046 & 97,837  \\
Mortality        & 334,515 & 41,814 & 41,815 & 51,018 & 97,586 & 566,748 \\
\bottomrule
\end{tabular}
\label{tab:note_counts}
\end{table}

\begin{table}[ht]
\centering
\caption{Distinct patient counts for each ReMedE task across five splits.}
\begin{tabular}{lcccccc}
\toprule
\textbf{Task} & \textbf{Train} & \textbf{Val} & \textbf{Test} & \textbf{Temporal Test 2021} & \textbf{ Temporal Test 2024} & \textbf{Total} \\
\midrule
Readmission       & 269,140 & 42,692 & 42,603 & 46,003 & 78,453  & 421,429\\
CCI               & 188,298 & 30,085 & 30,098 & 251,804 & 64,873 & 306,741\\
Length of Stay    & 248,486 & 39,304 & 39,331 & 43,358 & 78,453  & 395,991\\
Insurance Denial  & 39,422  & 2,319  & 2,313  & 9,037  & 37,821  & 87,974\\
Mortality         & 248,674 & 39,317 & 39,357 & 43,358 & 78,453  & 395,991\\
\bottomrule
\end{tabular}
\label{tab:distinct_patients}
\end{table}


\begin{table}[ht]
\centering
\caption{Readmission label ratios by split (Total notes = 604{,}326; total words = 607{,}877{,}177).}
\begin{tabular}{lcc}
\toprule
\textbf{Split} & \textbf{Not Readmitted} & \textbf{Readmitted within 30 days} \\
\midrule
Train               & 0.891495 & 0.108505 \\
Val                 & 0.891944 & 0.108056 \\
Test                & 0.893293 & 0.106707 \\
Temporal Test (2021)& 0.888456 & 0.111544 \\
New Temporal (2024) & 0.887115 & 0.112885 \\
\bottomrule
\end{tabular}
\label{tab:readmission_ratios}
\end{table}

\begin{table}[ht]
\centering
\caption{In-hospital mortality label ratios by split (Total notes = 566{,}748; total words = 608{,}603{,}182).}
\begin{tabular}{lcc}
\toprule
\textbf{Split} & \textbf{Survived} & \textbf{Died} \\
\midrule
Train               & 0.981161 & 0.018839 \\
Val                 & 0.981250 & 0.018750 \\
Test                & 0.980916 & 0.019084 \\
Temporal Test (2021)& 0.980693 & 0.019307 \\
New Temporal (2024) & 0.982241 & 0.017759 \\
\bottomrule
\end{tabular}
\label{tab:mortality_ratios}
\end{table}

\begin{table}[ht]
\centering
\caption{Length of stay label ratios by split (Total notes = 566{,}748; total words = 608{,}603{,}182).}
\begin{tabular}{lcccc}
\toprule
\textbf{Split} & \textbf{0--2 days} & \textbf{3 days} & \textbf{4--5 days} & \textbf{>5 days} \\
\midrule
Train               & 0.417306 & 0.176575 & 0.165888 & 0.240231 \\
Val                 & 0.415722 & 0.180562 & 0.164490 & 0.239226 \\
Test                & 0.414851 & 0.176611 & 0.167452 & 0.241086 \\
Temporal Test (2021)& 0.418597 & 0.153201 & 0.164805 & 0.263397 \\
New Temporal (2024) & 0.456418 & 0.144529 & 0.159757 & 0.239297 \\
\bottomrule
\end{tabular}
\label{tab:los_ratios}
\end{table}

\begin{table}[ht]
\centering
\caption{Charlson Comorbidity Index (CCI) label ratios by split (Total notes = 443{,}915; total words = 524{,}739{,}038).}
\begin{tabular}{lccccc}
\toprule
\textbf{Split} & \textbf{Score 0} & \textbf{Score 1--2} & \textbf{Score 3--4} & \textbf{Score 5--7} & \textbf{Score >7} \\
\midrule
Train               & 0.694681 & 0.224840 & 0.054251 & 0.025324 & 0.000904 \\
Val                 & 0.689169 & 0.229547 & 0.055166 & 0.025526 & 0.000592 \\
Test                & 0.693502 & 0.226866 & 0.053327 & 0.025370 & 0.000935 \\
Temporal Test (2021)& 0.685811 & 0.228327 & 0.059164 & 0.025821 & 0.000878 \\
New Temporal (2024) & 0.684043 & 0.217355 & 0.067513 & 0.029370 & 0.001717 \\
\bottomrule
\end{tabular}
\label{tab:cci_ratios}
\end{table}

\begin{table}[ht]
\centering
\caption{Insurance denial label ratios by split (Total notes = 97{,}837; total words = 89{,}147{,}715).}
\begin{tabular}{lcc}
\toprule
\textbf{Split} & \textbf{Approved (0)} & \textbf{Denied (1)} \\
\midrule
Train               & 0.877850 & 0.122150 \\
Val                 & 0.867097 & 0.132903 \\
Test                & 0.873978 & 0.126022 \\
Temporal Test (2021)& 0.879880 & 0.120120 \\
New Temporal (2024) & 0.861033 & 0.138967 \\
\bottomrule
\end{tabular}
\label{tab:insurance_ratios}
\end{table}

\clearpage
\newpage

\section{Transfer Pattern of \textsc{Llama 3.2 1b} v.s. \textsc{Lang1 1B}}\label{app:llama_3_2_transfer_vs_lang1}

\begin{figure}[ht!]
    \centering
        \centering
    \begin{subfigure}{0.48\textwidth}
        \includegraphics[width=\linewidth]{figures/main/transfer/transfer_learning_heatmap_no_intermediate_Lavender.pdf}
        \caption{\textsc{Lang1-1B}'s transfer heatmap.  }\label{fig:lang1_transfer}
     \end{subfigure}
    \begin{subfigure}{0.48\textwidth}
        \includegraphics[width=\linewidth]{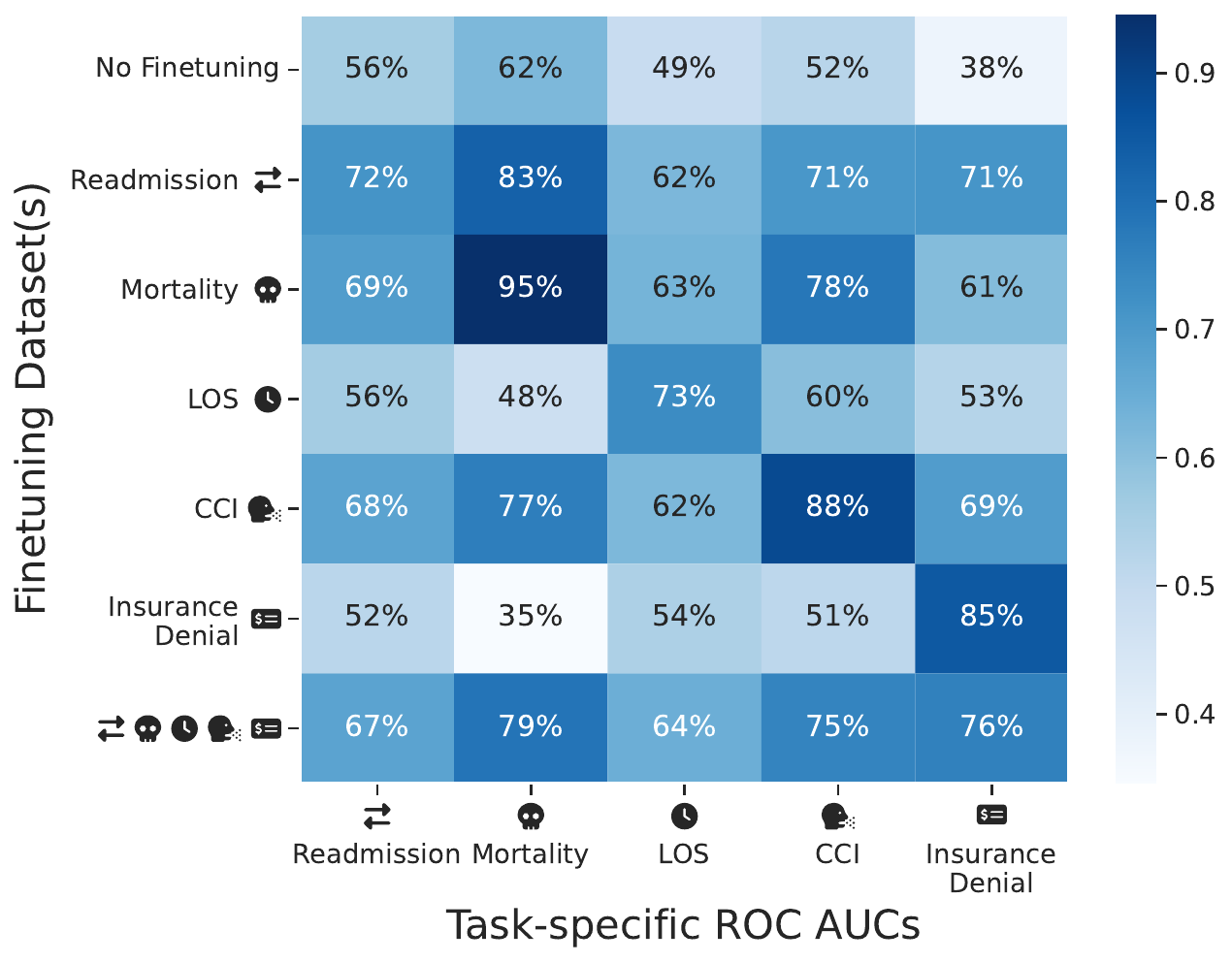}
        \caption{\textsc{Llama 3.2 1B}'s transfer heatmap. 
        }\label{fig:small_llama_transfer}
    \end{subfigure}
    \caption{\textbf{\textsc{Lang1-1B} and \textsc{Llama 3.2 1B} transfers differently.} The heatmap shows the two different base models' performance when finetuned on a subset of \textsc{ReMedE} tasks ($y$ axis) and evaluated on all five \textsc{ReMedE} tasks ($x$ axis). Overall, \textsc{Lang1-1B} has higher per-task and joint performance, and shows a different transfer pattern than \textsc{Llama 3.2 1B}.}
    \label{fig:transfer-learning-heatmap}
\end{figure}

\vspace{-20pt}
\paragraph{\textsc{Llama-3.2-1B} has overall worse performance than \textsc{Lang1-1B}} Compared to \autoref{fig:lang1_transfer}, \autoref{fig:small_llama_transfer} has worse single-task~(diagonal) and joint-task~(last row) performance, suggesting that the specific transfer pattern could be highly model specific.

\paragraph{Both models exhibit some similar patterns} (i) finetuning on readmission (second row) boosts performance on the other four tasks, and (ii) transfer could be asymmetric, \textit{i.e.}, mortality helps LOS (third row, third column) but LOS does not help mortality (fourth row, second column), which can be explained using domain knowledge (\autoref{app:asymmetry}). 

\paragraph{\textsc{Lang1-1B} and \textsc{Llama-3.2-1B} also have different patterns} (i) joint finetuning (last row) helps \textsc{Lang1-1B} but hurts \textsc{Llama-3.2-1B}, and (ii) finetuning on insurance denial (fourth row) lowers \textsc{Lang1-1B}'s LOS performance (third column) while improving it for \textsc{Llama-3.2-1B}. These results suggest that \textbf{instruction finetuning enables cross-task transfer, though the specific transfer patterns depend on model pretraining}.

\newpage

\section{Pretraining ablations}\label{sec:pretrain_ablation}



\begin{figure}[ht!]
    \centering
    \begin{subfigure}[t]{0.65\linewidth}
        \centering
        \includegraphics[width=\linewidth]{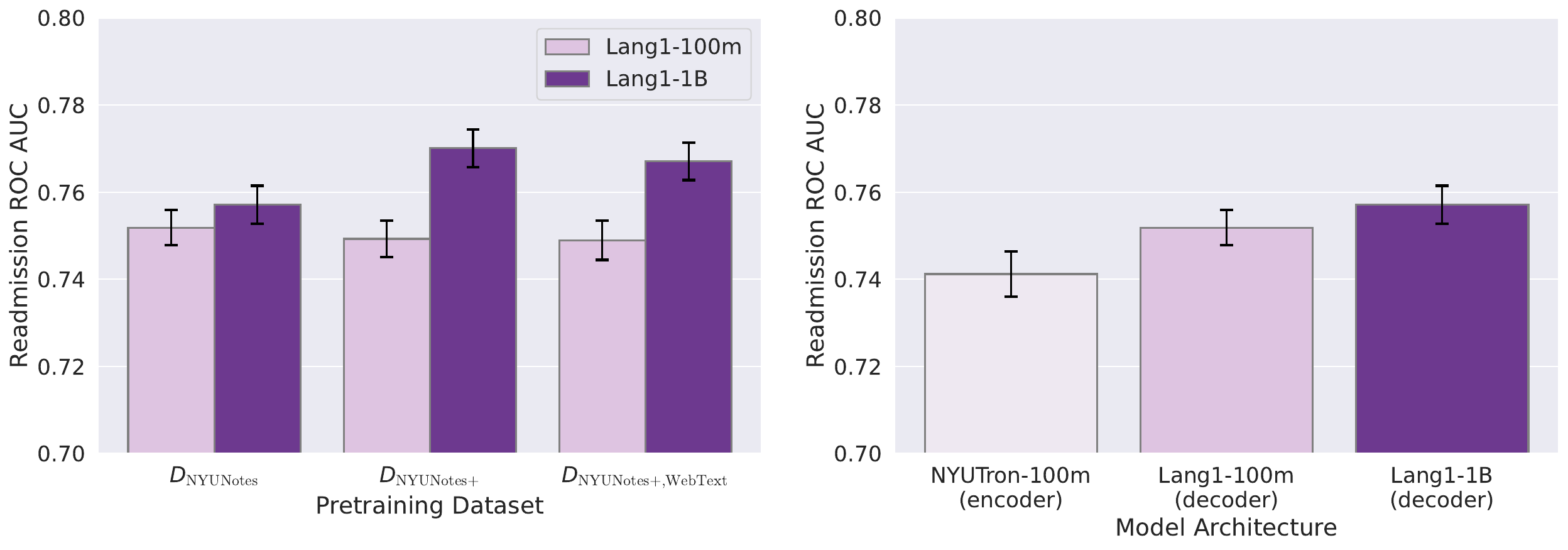}
        \caption{Readmission prediction}
        \label{fig:pretrain_ablation_readmission}
    \end{subfigure}

      \begin{subfigure}[t]{0.48\linewidth}
        \centering
        \includegraphics[width=\linewidth]{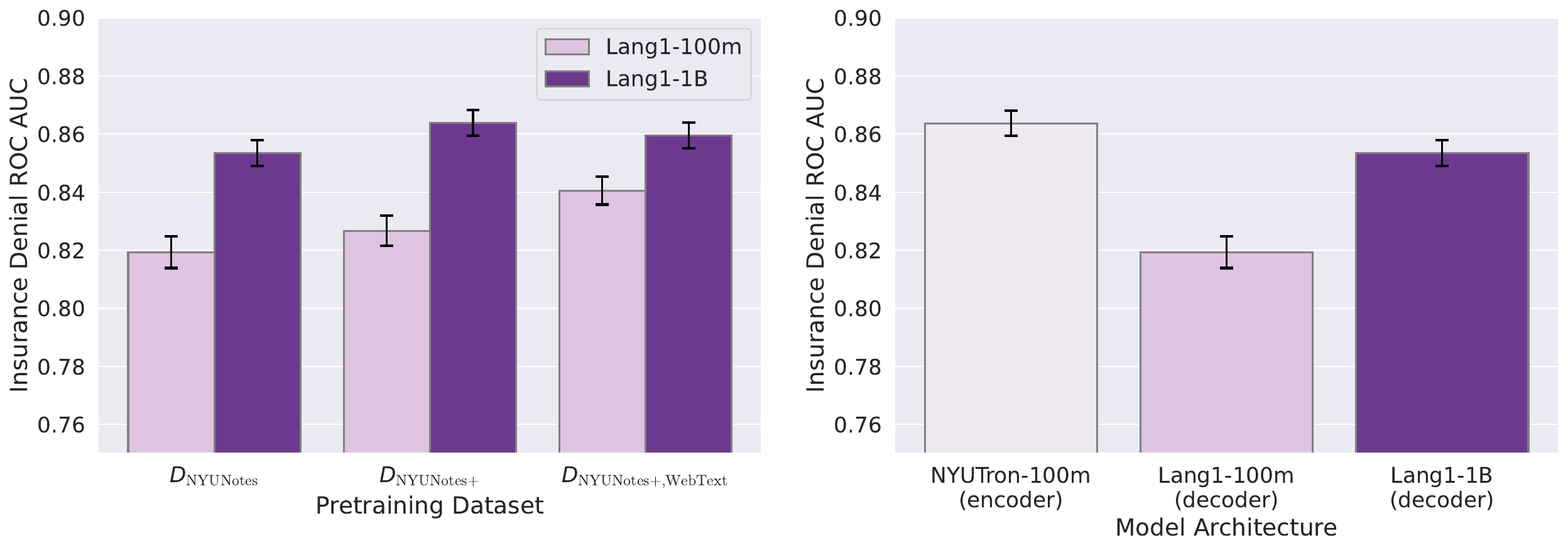}
        \caption{Insurance denial prediction}\label{fig:pretrain_ablation_insurance}
\end{subfigure}
\hfill
      \begin{subfigure}[t]{0.48\linewidth}
        \centering
        \includegraphics[width=\linewidth]{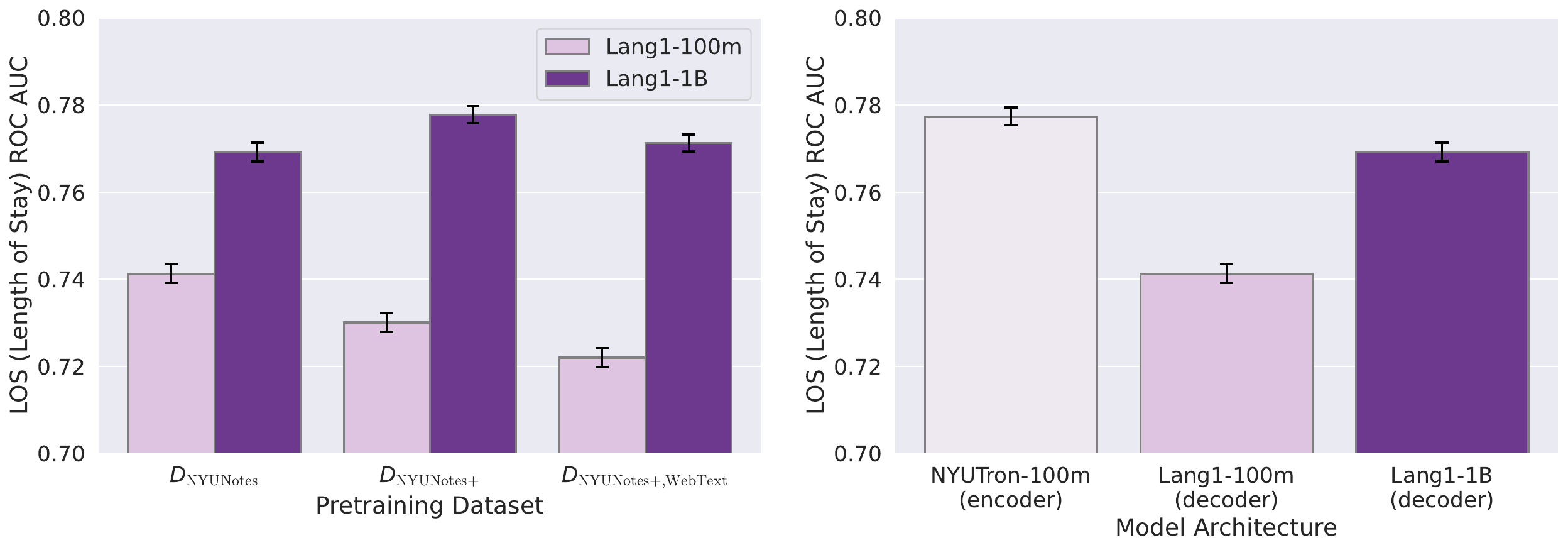}
        \caption{LOS prediction}\label{fig:pretrain_ablation_los}
\end{subfigure}
    \caption{\textbf{Pretraining ablations.} Error bars indicating the 95\% confidence interval. (a) Larger models trained on more clinical data has better performance, and mixing web texts does not hurt much. (b,c) Similar patterns are observed for insurance denial and LOS prediction.}
    \label{fig:combined_robustness}
\end{figure}

\autoref{fig:combined_robustness} presents pretraining ablation results evaluated on 2024 readmission temporal test set. For pretraining, we control for the model architecture (encoder v.s. decoder), model size (100M v.s. 1B), and pretrain data (NYUNotes, NYUNotes$+$, NYUNotes$+$ and web texts). For finetuning, we evaluated on three clinical predictive tasks: readmission prediction, insurance denial prediction, and LOS prediction. These three tasks were chosen for their distinct transfer pattern in \autoref{fig:lang1_transfer}.

\paragraph{Training larger models on more recent clinical data improves temporal robustness} \autoref{fig:pretrain_ablation_readmission} shows the ablation results for readmission prediction. On the left, holding the model architecture fixed as a decoder, we vary pretraining data. Compared to models trained only on EHR from 2013 to 2021 ($\mathcal{D}_\text{NYUNotes}$), adding more recent clinical data ($\mathcal{D}_\text{NYUNotes+}$, spanning 2003 to 2023) and further mixing in general-domain SlimPajama ($\mathcal{D}_\text{NYUNotes+,WebText}$) improves performance for the 1B model but not the 100M model, suggesting that larger models are better able to leverage additional clinical data. \textbf{This also justifies our choice of equally mixing EHR texts and web texts}, since adding web texts does not significantly hurt the downstream clinical task performance while instilling general-purpose knowledge. On the right, holding the pretraining data fixed to $\mathcal{D}_\text{NYUNotes}$, varying model architecture shows that scaling to 1B helps. We observe similar patterns for insurance denial prediction (\autoref{fig:pretrain_ablation_insurance}) and LOS prediction (\autoref{fig:pretrain_ablation_los}). 

\newpage
\section{Calibration Plot}\label{sec:cali}

Calibration curves are calculated using sklearn package~\citep{Pedregosa2012-nl} with n=15 bins. Expected calibration error (ECE) is calculated with n=15 bins using the torchmetrics library~\citep{Detlefsen2022-mb}.

\begin{figure}[htp!]
     \centering
     \begin{subfigure}[b]{\textwidth}
         \centering
         \includegraphics[width=\textwidth]{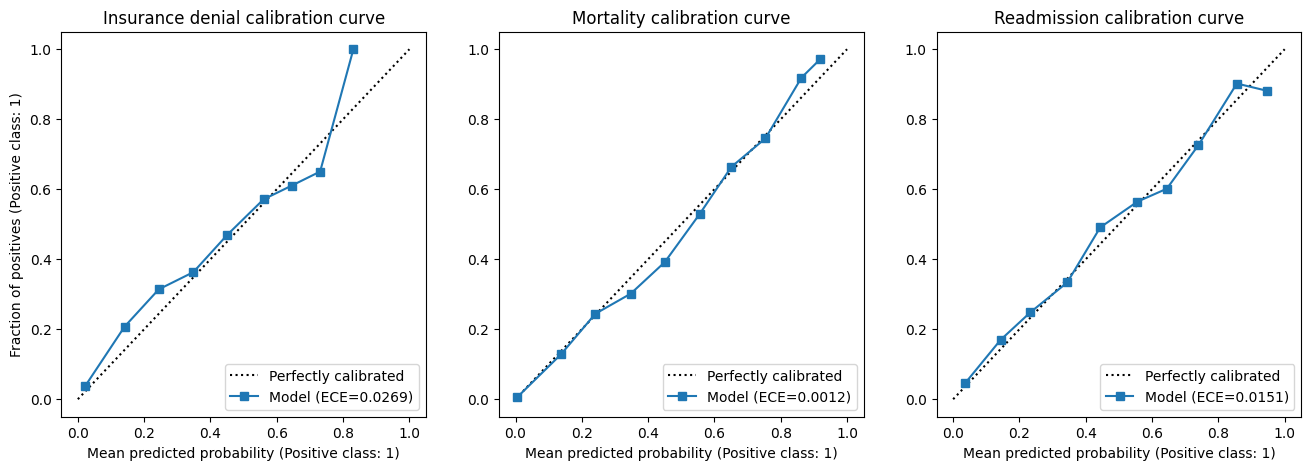}
         \caption{Calibration plots for single-task models}
         \label{fig:cali_single}
     \end{subfigure}
     \hfill
     \begin{subfigure}[b]{\textwidth}
         \centering
         \includegraphics[width=\textwidth]{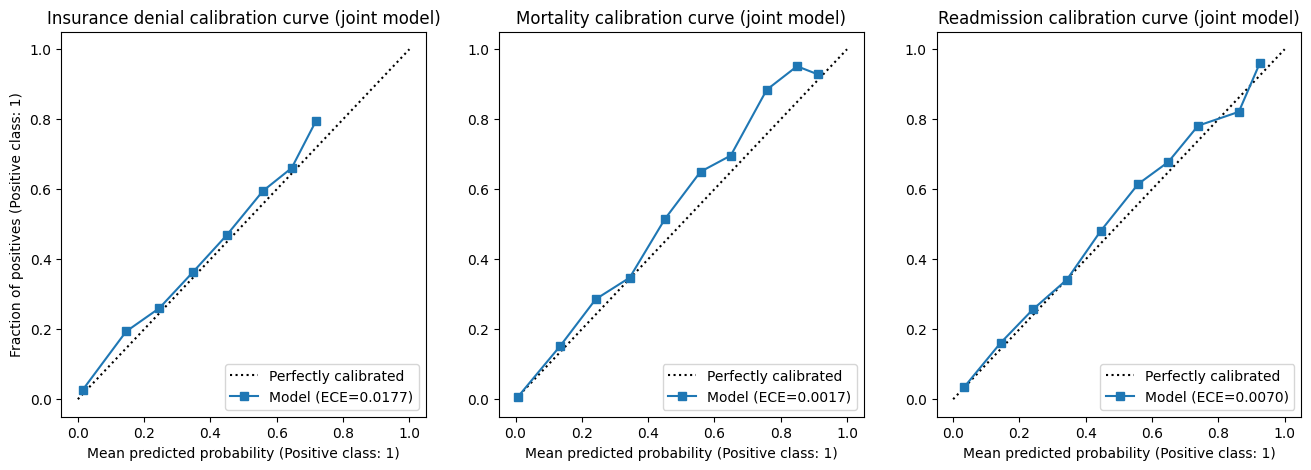}
         \caption{Calibration plots for joint model}
         \label{fig:cali_joint}
     \end{subfigure}
       \caption{Calibration plot shows that both single task finetuned Lang1 and joint finetuned Lang1 are well calibrated.}
\end{figure}

\newpage

\section{Asymmetry of Transfer between Mortality and LOS}\label{app:asymmetry}

Our medical collaborators provided an explanation for the asymmetric transfer between Mortality and LOS (Length of Stay) we observe for both \textsc{Lang1-1B} and \textsc{Llama-3.2-1B}. If a patient died, they either stayed for a short time (very sick and died immediately) or a long time (doctors failed to save them after a long time). On the other hand, if a patient stayed for a long time, they either survived, or they died after doctors' attempts. This asymmetry in conditional probability could help explain why the mortality transfer to LOS, but not vice versa. 

\begin{figure}[htp!]
  \centering
  \begin{subfigure}[t]{0.48\textwidth}
    \centering
    \includegraphics[width=\linewidth]{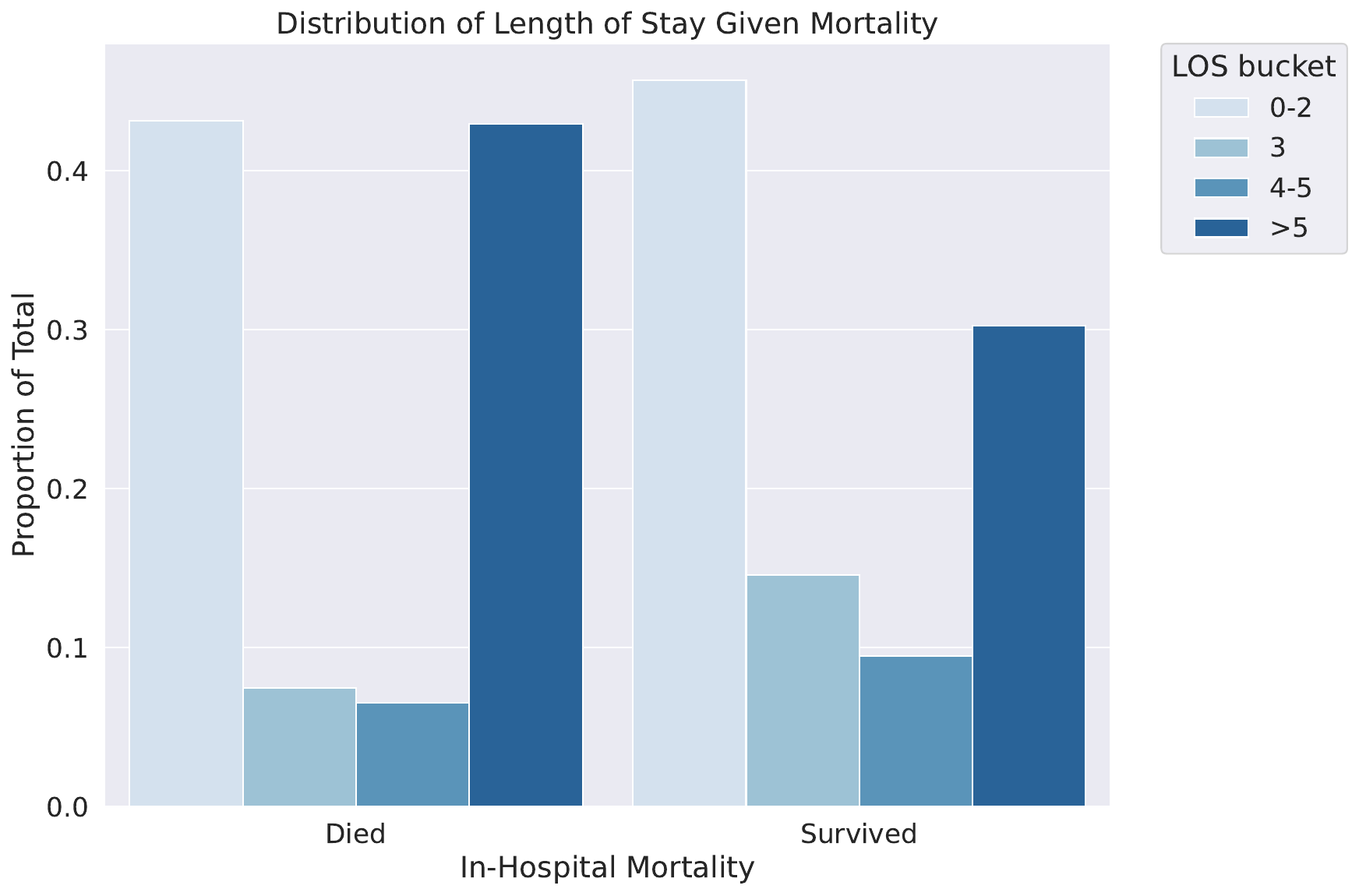}
    \caption{Probabilities of LOS given in-hospital mortality}
    \label{fig:cond_given_mort}
  \end{subfigure}\hfill
  \begin{subfigure}[t]{0.48\textwidth}
    \centering
    \includegraphics[width=\linewidth]{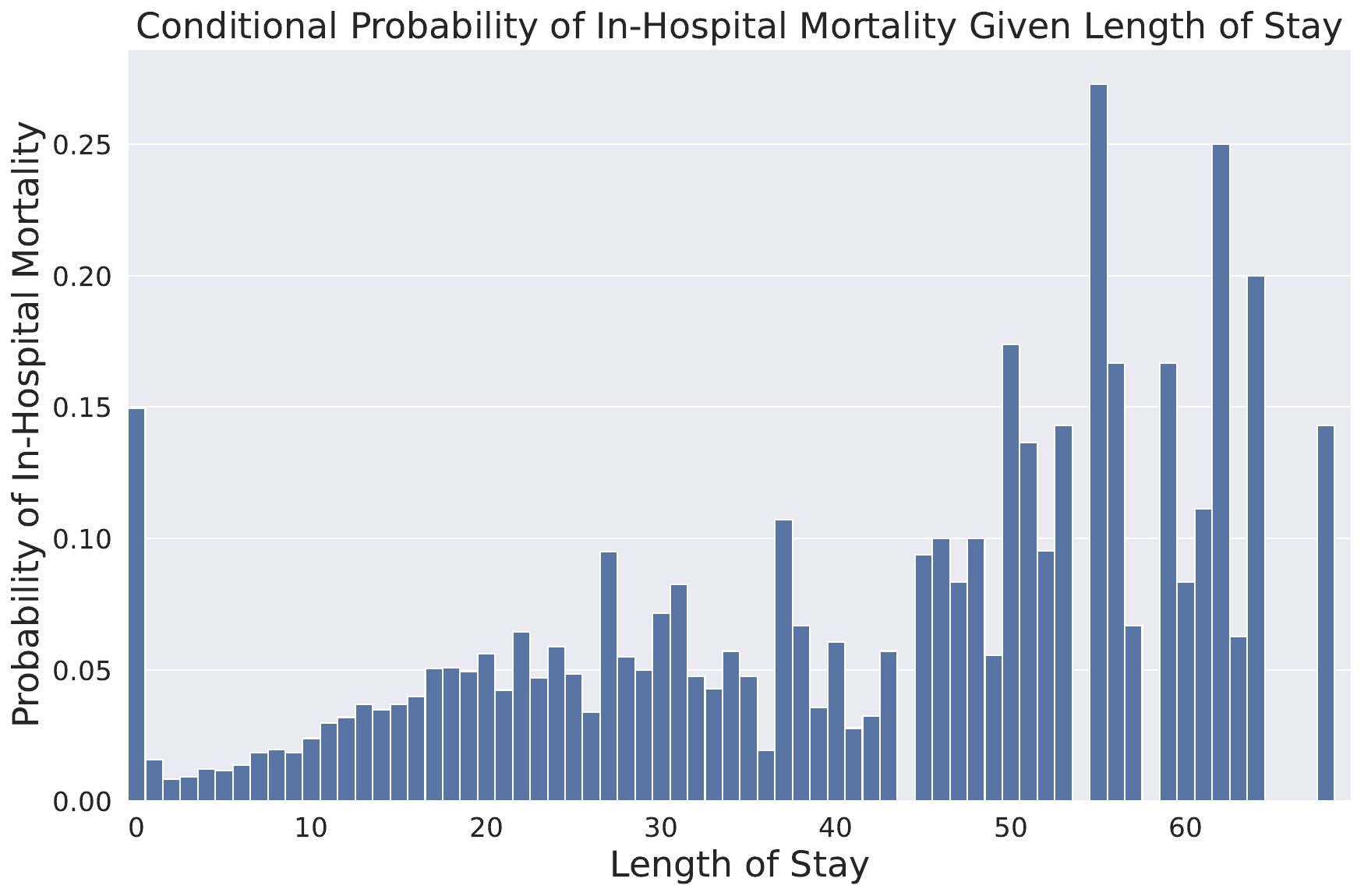}
    \caption{Probabilities of in-hospital mortality given LOS}
    \label{fig:cond_given_los} 
  \end{subfigure}
  \caption{Analysis of conditional probabilities of LOS (length of stay) and in-hospital mortality could help explain the assymetry of transfer.}
  \label{fig:cond_prob}
\end{figure}

This explanation is corroborated by analysis of the conditional probabilities. \autoref{fig:cond_given_mort} shows that patient who died are ore likely to stay for 0-2 days or >5 days compared to patient who survived. \ref{fig:cond_given_los} shows that patient who stay for a long time have similar mortality risks as patient who stayed for 0 days.


\newpage

\section{Detailed statistics of NYU Notes+}\label{sec:nyunotes+_stats}

\begin{figure}[htbp]
    \centering
    \begin{subfigure}{0.8\textwidth}
        \includegraphics[width=\linewidth]{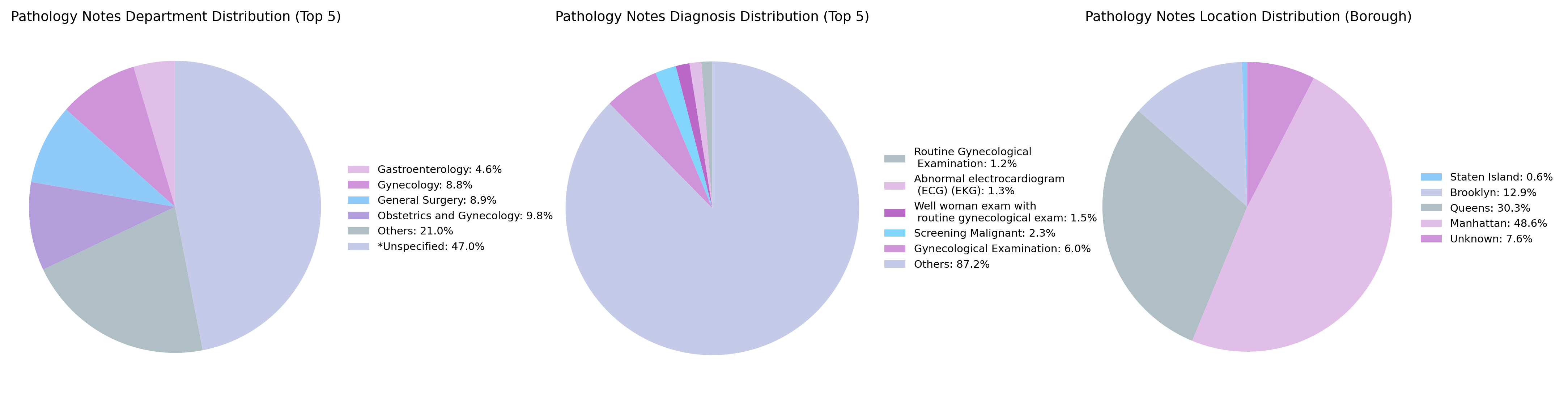}
        \caption{\textbf{Pathology Notes.} Among pathology notes with specified departments, OB/GYN (obstetrics and gynecology) has the highest percentage (9.8\%). The most common specified diagnosis is gynecological exams (6\%). Nearly half (48.6\%) of the pathology notes are from Manhattan borough.}
    \end{subfigure}

    \vspace{0.5em} 

    \begin{subfigure}{0.8\textwidth}
        \includegraphics[width=\linewidth]{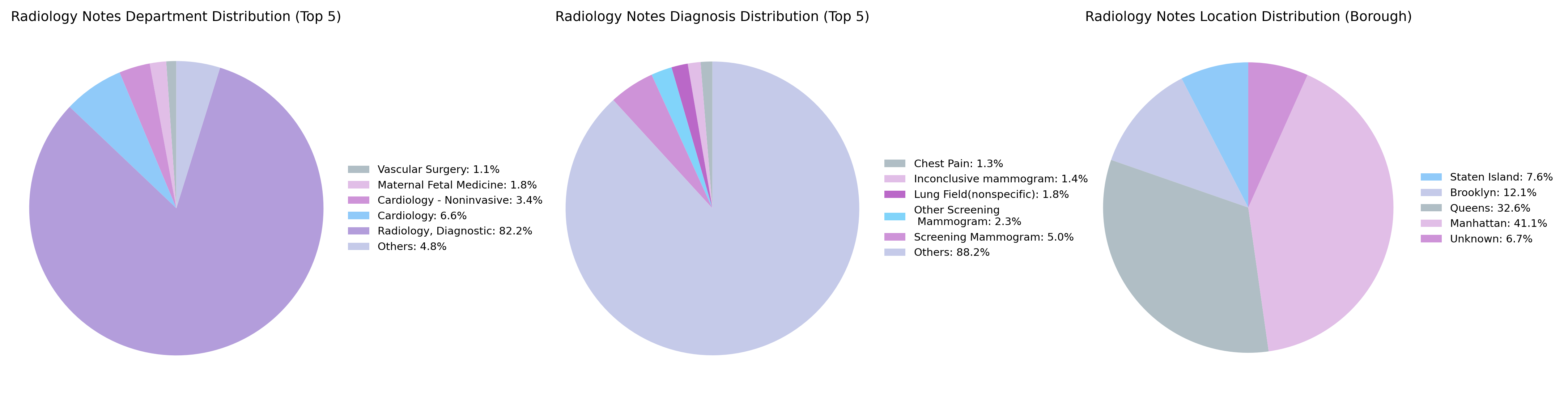}
        \caption{\textbf{Radiology Notes.} Most of the radiology notes are from diagnostic radiology (82.2\%), with screening mammograms being the most common specified diagnosis (5\%). The two most common boroughs are Manhattan (41\%) and Queens (32.6\%).}
    \end{subfigure}

    \vspace{0.5em}

    \begin{subfigure}{0.8\textwidth}
        \hspace{-0.02\textwidth}
        \includegraphics[width=\linewidth]{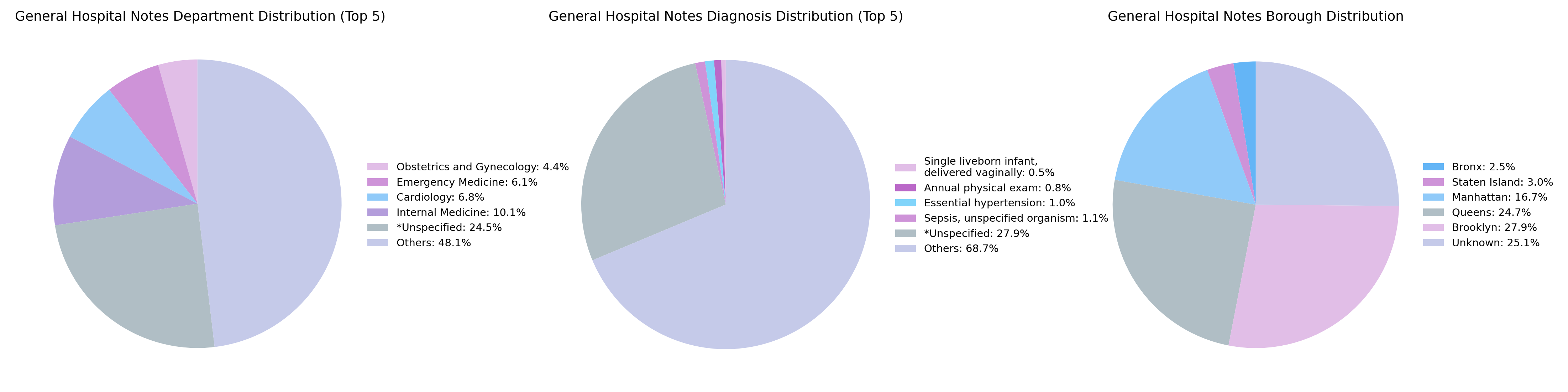}
        \caption{\textbf{Hospital Notes.} Among notes with specified departments, most are from internal medicine department (10.1\%), with common diagnoses including sepsis (1.1\%) and hypertension (1\%). The majority of notes are from Brooklyn (27.9\%) and Queens (24.7\%) borough.}
    \end{subfigure}

    \caption{Top five department, diagnosis and borough of NYU Notes+}
    \label{fig:nyunotes+_stats}
\end{figure}

We analyzed the top five clinical departments, diagnosis and borough for pathology notes, radiology notes and hospital notes. See Figure~\ref{fig:nyunotes+_stats}.

\newpage

\section{Prompts of ReMedE Tasks}\label{sec:detailed_prompts}

We constructed prompts to create task-specific questions and answer options from the labeled finetuning notes:

\begin{itemize} 
    \item \textbf{Readmission} 
    Question: Given the above discharge note of the patient, will the patient be readmitted to the hospital within 30 days of discharge? $\backslash$n A. no $\backslash$n B. yes $\backslash$n Answer:

    \item \textbf{In-Hospital Mortality} 
    Question: Given the above admission note of the patient, will the patient die during the hospital admission? $\backslash$n A. no $\backslash$n B. yes $\backslash$n Answer:

    \item \textbf{Charlson Comorbidity Index}  
    Question: Given the above admission note of the patient, what's the Charlson Comorbidity Index of the patient? $\backslash$n A. score 0 $\backslash$n B. score 1 to 2 $\backslash$n C. score 3 to 4 $\backslash$n D. score 4 to 7 $\backslash$n E. score more than 7 $\backslash$n Answer:

    \item \textbf{Insurance Denial} 
    Question: Given the above discharge note of the patient, will the insurance claim of the patient be denied? $\backslash$n A. no $\backslash$n B. yes $\backslash$n Answer:

    \item \textbf{Length of Stay} 
    Question: Given the above admission note of the patient, how long will the patient stay at the hospital? $\backslash$n A. 0 to 2 days $\backslash$n B. 3 days $\backslash$n C. 4 to 5 days $\backslash$n D. more than 5 days $\backslash$n Answer:
\end{itemize}

\section{LoRa Finetuning for Llama-3-70b}\label{app:lora}

To efficiently train Deepseek-R1-distill-Llama-3-70b on 1 node of 8 H100s, we used Low-Rank Approximation (LoRA) finetuning. LoRA reduces trainable parameters by inserting trainable rank decomposition matrices into transformer layers while freezing the pretrained weights. 

In our configuration, we enabled LoRA adapters on the query and value projections of the attention mechanism, with rank $r=8$. We set the LoRA scaling factor $\alpha=16$ and applied a dropout rate of 0.05. Other components, such as the key, MLP, and projection layers, were left frozen. 

\section{Token probability approximation for models without logprobs}\label{sec:prob_approx}

Unlike other generalist models, GPT-4o does not provide logprobs to prevent privacy attack. However, we need probabilities to reliably evaluate classification tasks. To approximate its probabilities, we sample 10 generations from GPT using temperature of 1 (to not reweight the LM's distribution), and count the number of occurrences for each multiple choice options. Then we normalize the counts to be probabilities for each option. For cost reasons, we limit the number of examples to be 1000, except for CCI (comorbidity imputation). 

CCI is a special case because its label distribution is very skewed, so we greedily evaluate on 10,000 samples instead. We binarize the greedy choice to be probability of 0 or 1. 

\newpage
\section{Stratified Evaluation}

\begin{figure}[ht]
\centering
\begin{subfigure}{0.3\columnwidth}
    \centering
    \includegraphics[width=\linewidth]{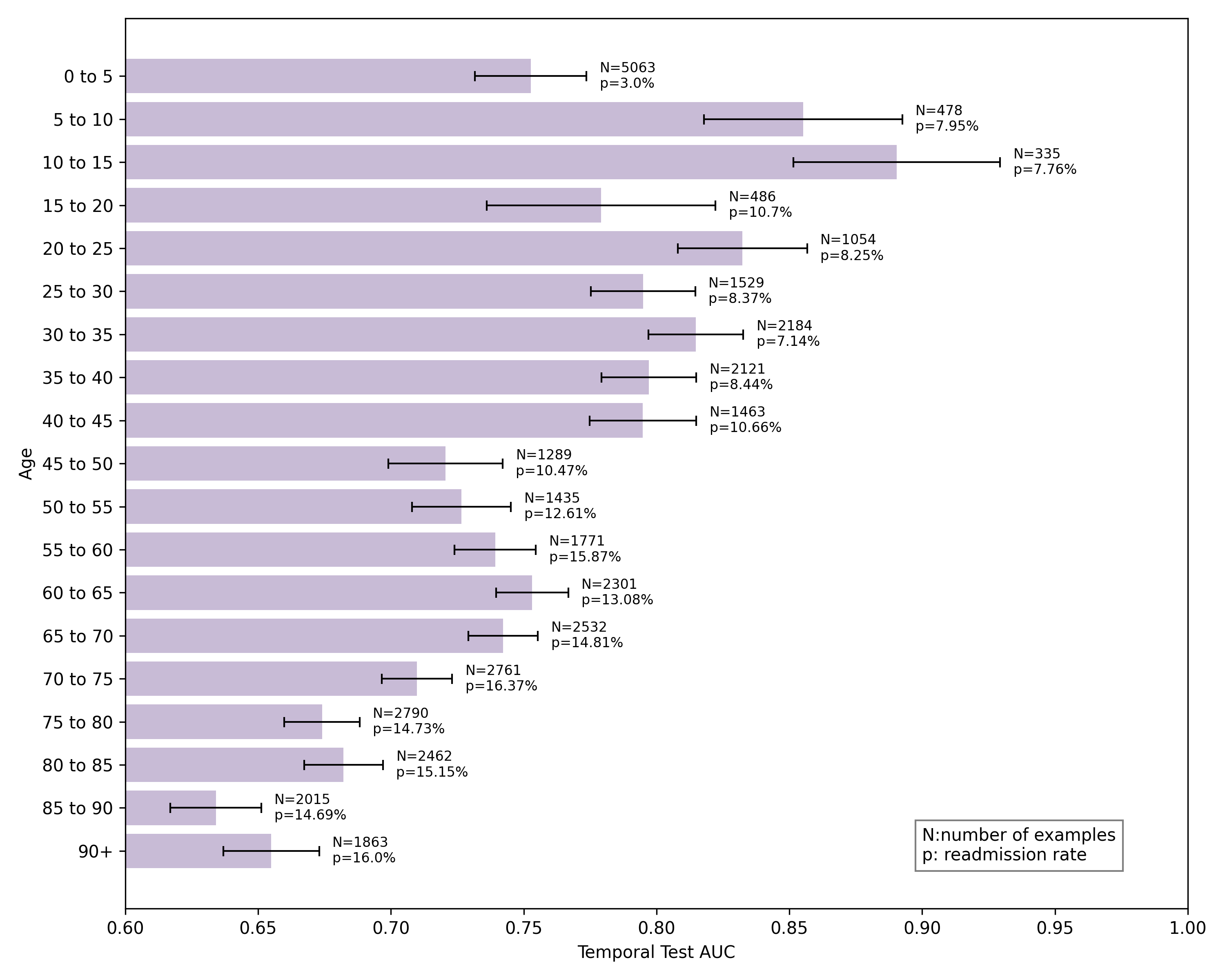}
    \subcaption{Age.}\label{fig:age}
\end{subfigure}
\hfill
\begin{subfigure}{0.3\columnwidth}
    \centering
    \includegraphics[width=\linewidth]{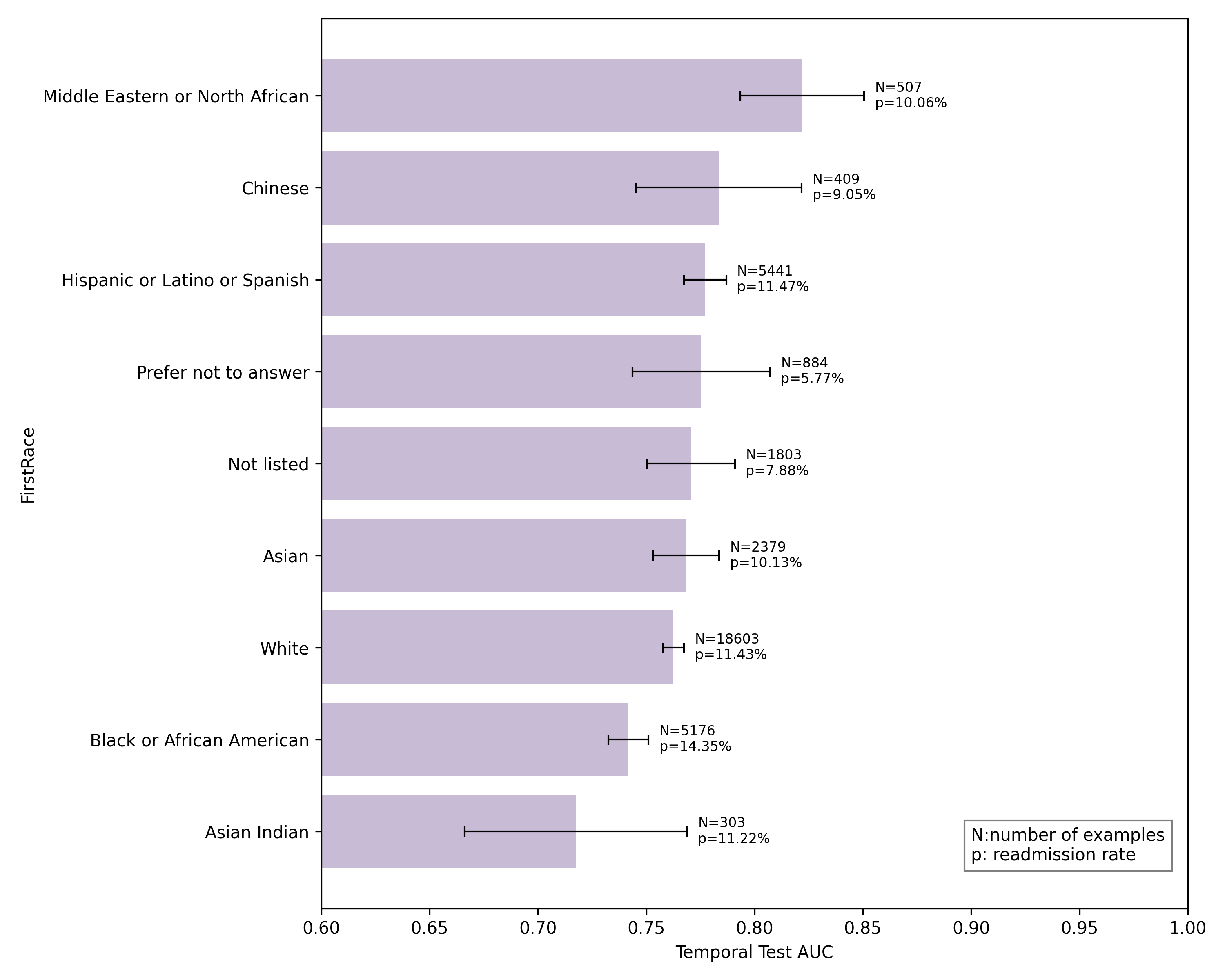}
    \subcaption{First Race.}
\end{subfigure}
\hfill
\begin{subfigure}{0.3\columnwidth}
    \centering
    \includegraphics[width=\linewidth]{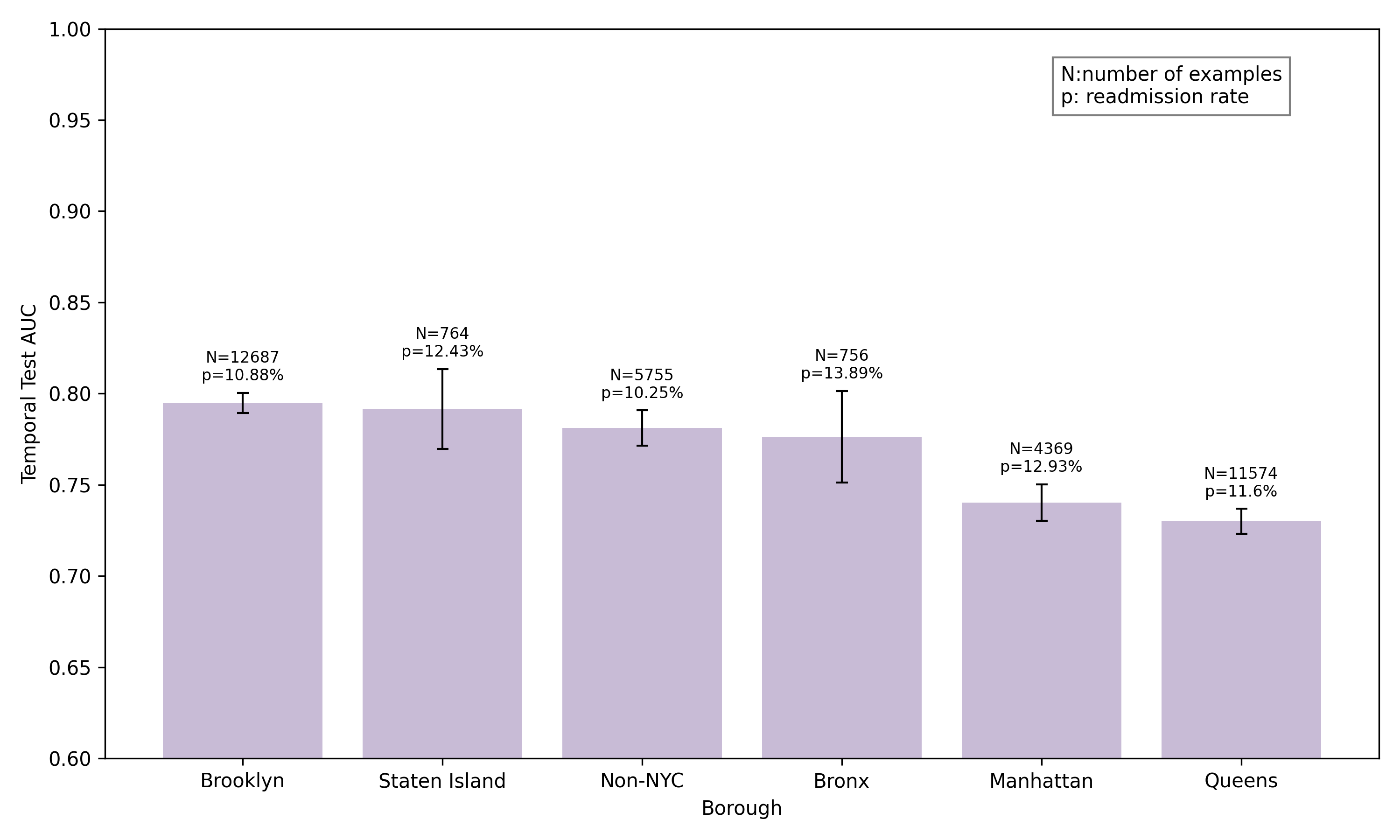}
    \subcaption{Borough.}
\end{subfigure}
\hfill
\begin{subfigure}{0.3\columnwidth}
    \centering
    \includegraphics[width=\linewidth]{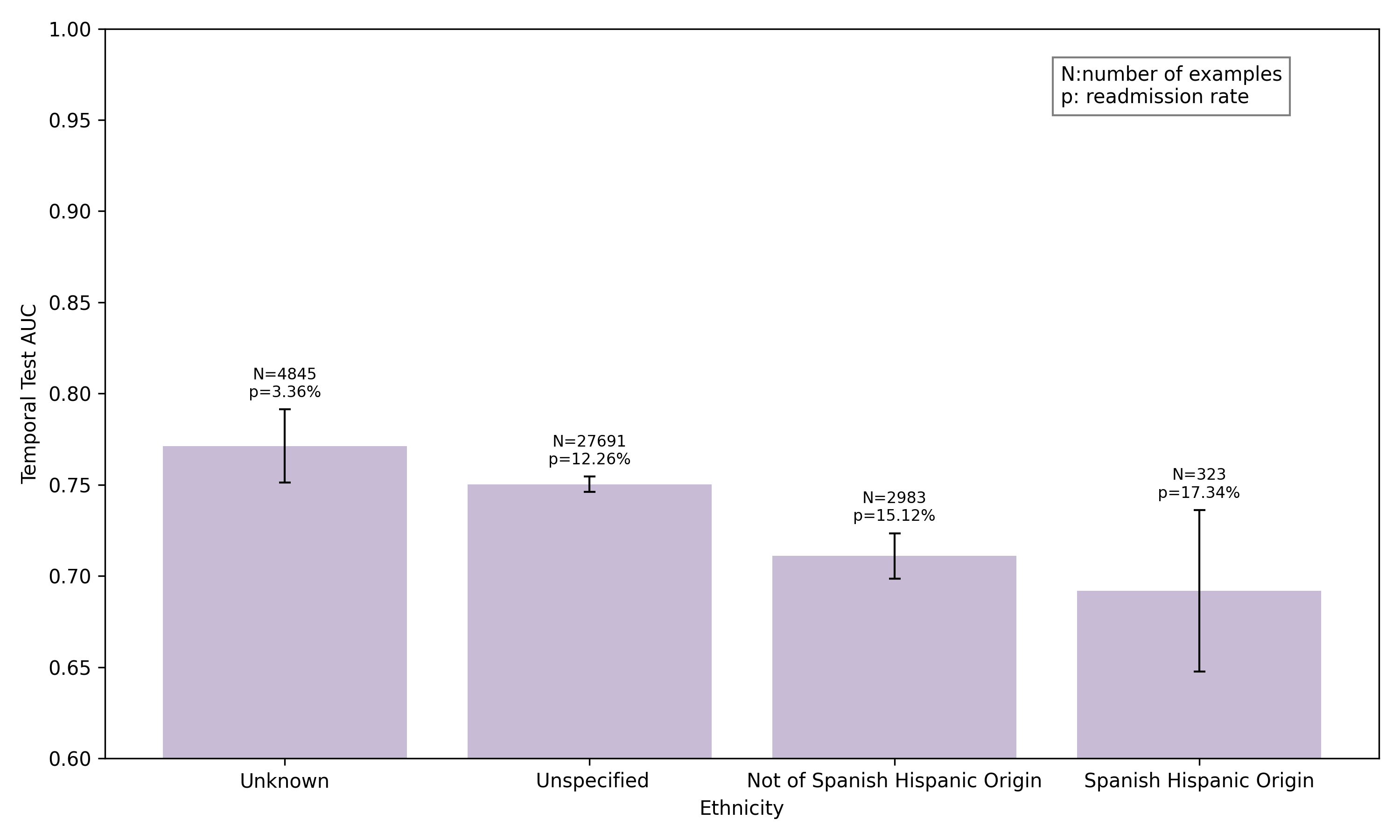}
    \subcaption{Ethnicity.}
\end{subfigure}
\hfill
\begin{subfigure}{0.3\columnwidth}
    \centering
    \includegraphics[width=\linewidth]{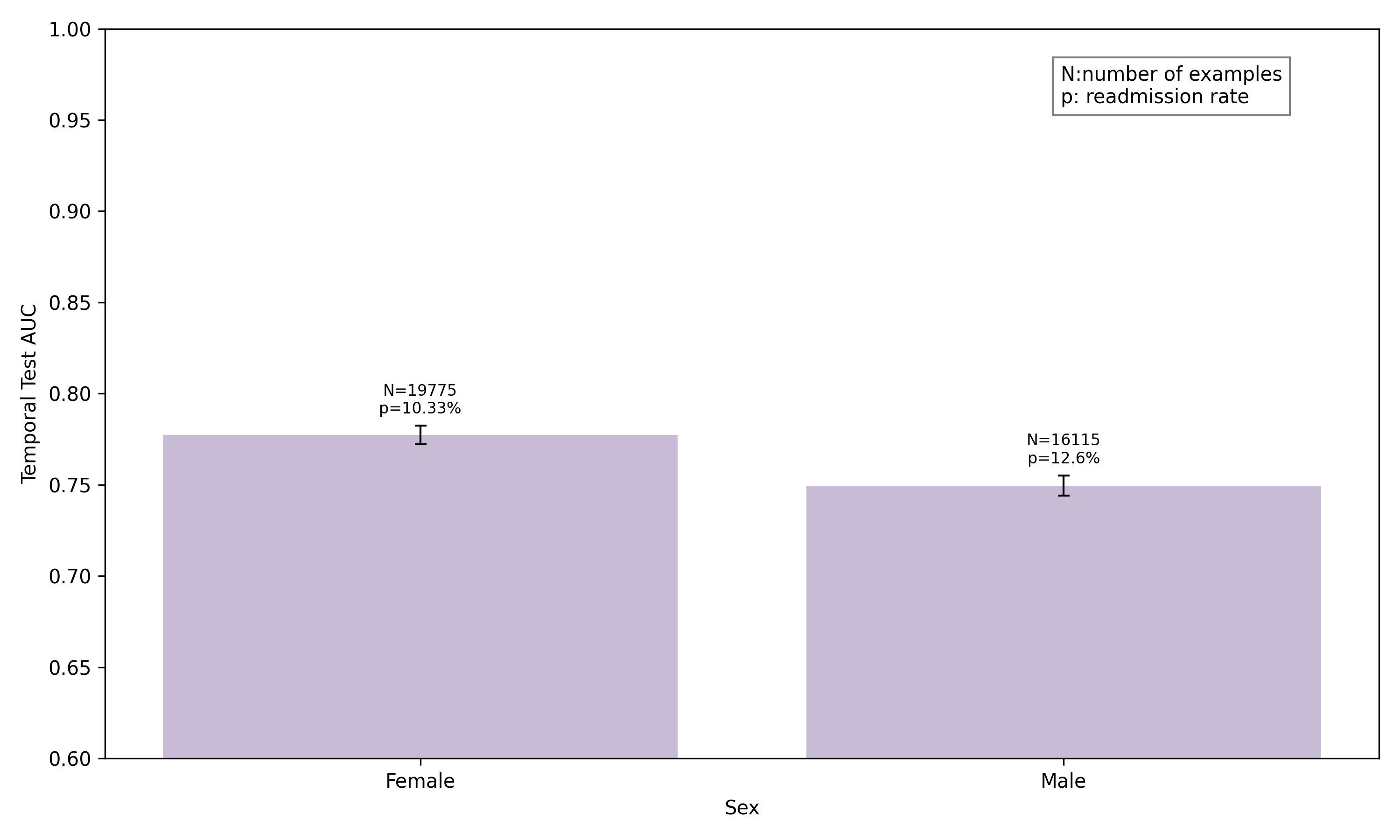}
    \subcaption{Sex.}
\end{subfigure}
\hfill
\begin{subfigure}{0.3\columnwidth}
    \centering
    \includegraphics[width=\linewidth]{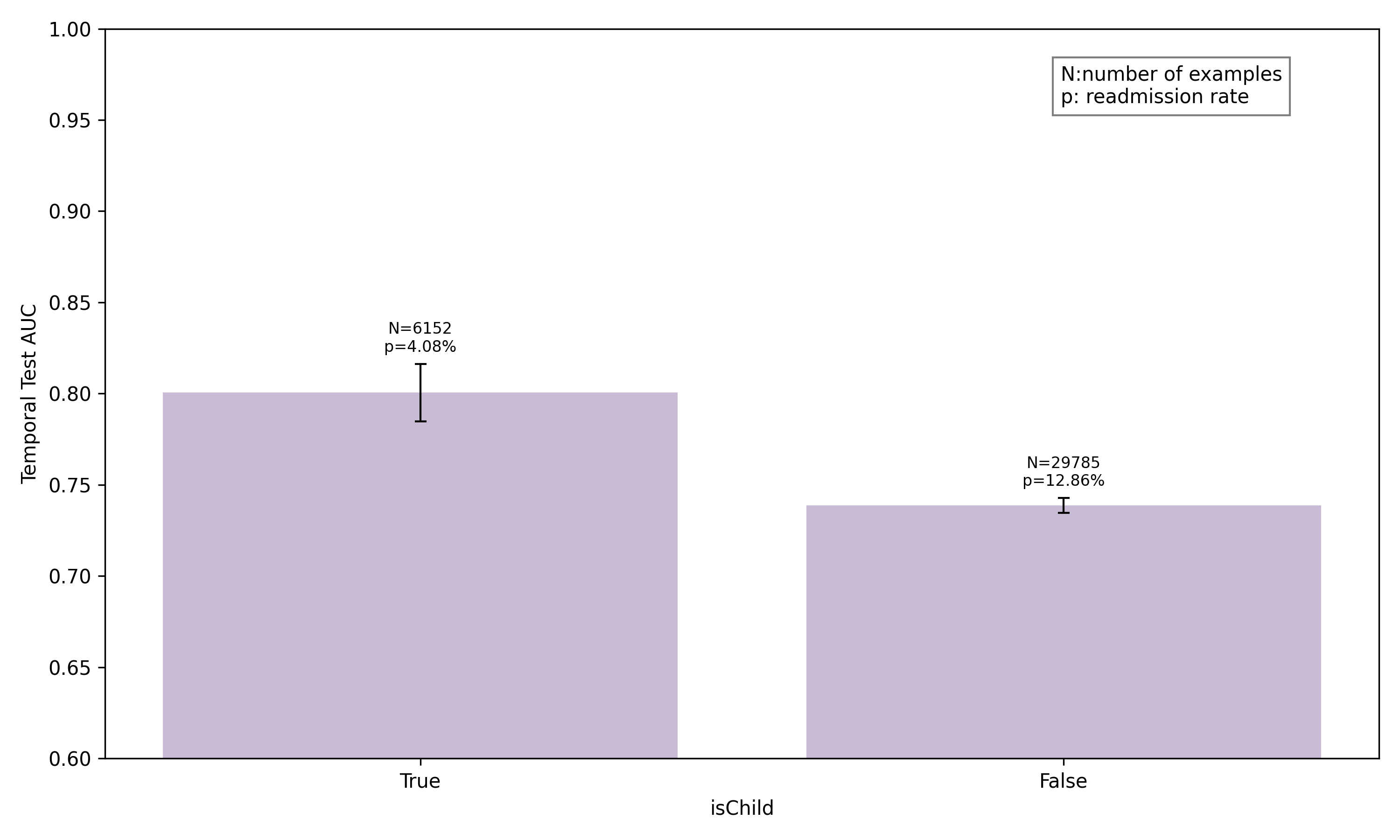}
    \subcaption{IsChild.}
\end{subfigure}

\caption{Stratified performance analysis on 2024 readmission temporal test set.}
\end{figure}

We performed stratified evaluation on readmission to evaluation the performance variation across different groups (age, first race, borough, ethnicity, sex, and whether the patients are children). Some groups are omitted because they have only one class due to small sample size. If we look at means only, for age (Figure \ref{fig:age}), patients between 10 to 15 has the best performance, and patients between 85 and 90 has the worst performance. For race, Middle Eastern or North African has the best performance, and Asian Indian has the worst performance. For borough, Brooklyn has the best performance and Queens has the worst performance. For ethnicity, Spanish Hispanic Origin is worst. For sex, female has better performance than male. Children's performance is better than adult and they also have a lower readmission rate.


\newpage

\section{Control for Patient Overlap}\label{app:patient_control}

\begin{figure}[htbp]
    \centering
    \begin{subfigure}{0.45\columnwidth}    
    \centering
    \includegraphics[width=\linewidth]{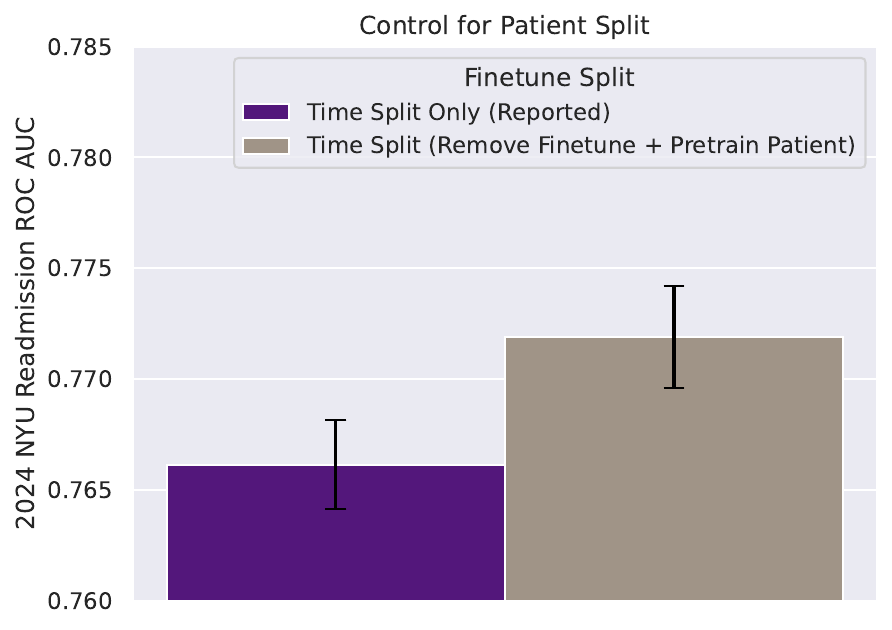}
    \subcaption{Readmission}\label{fig:patient_control_readmission}
    \end{subfigure}
    \hfill\\
       \begin{subfigure}{0.45\columnwidth}    
    \centering
    \includegraphics[width=\linewidth]{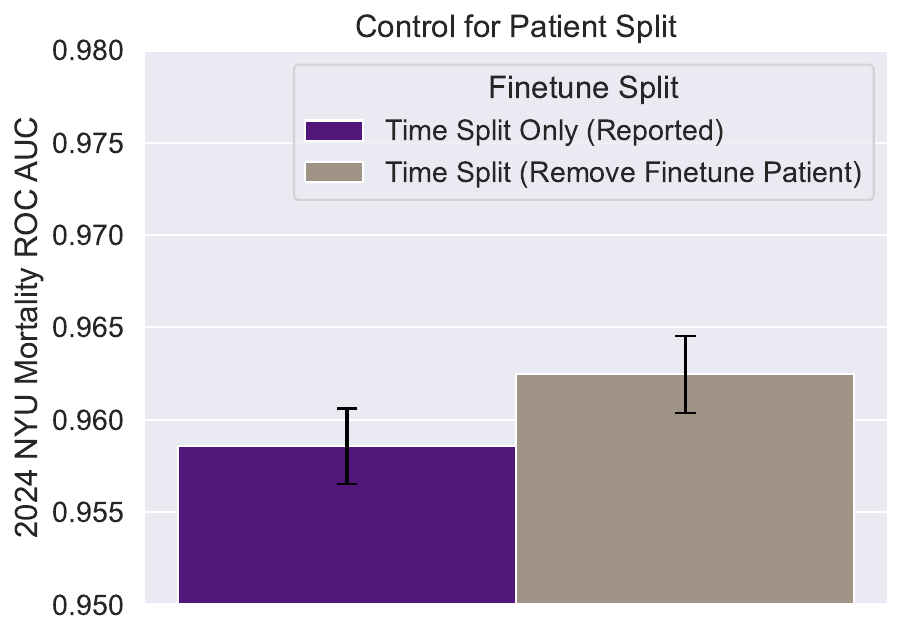}
    \subcaption{Mortality}\label{fig:patient_control_mortality}
    \end{subfigure}
          \begin{subfigure}{0.45\columnwidth}    
    \centering
    \includegraphics[width=\linewidth]{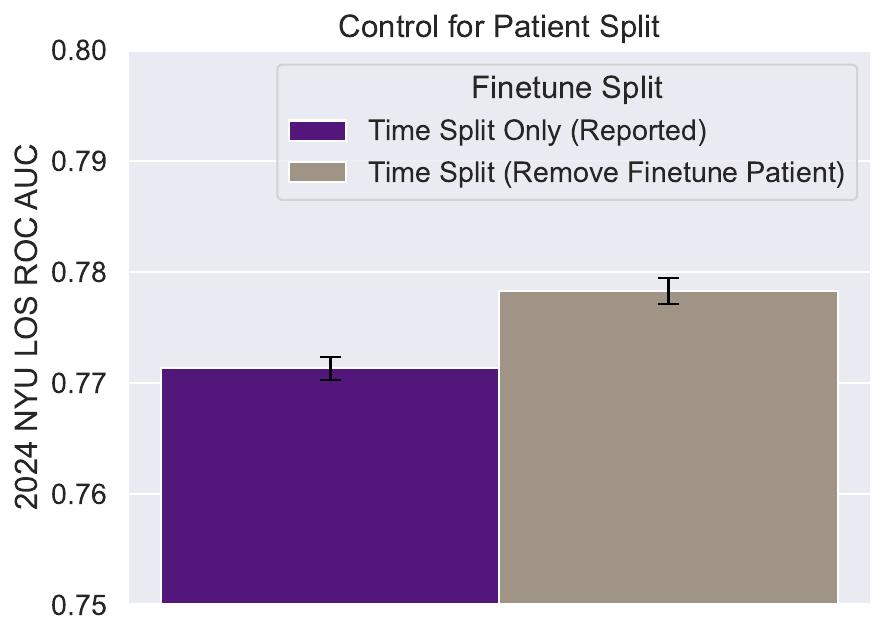}
    \subcaption{LOS}\label{fig:patient_control_los}
    \end{subfigure}
\caption{\textsc{Lang1-1B}'s finetuned performance on readmission, mortality and LOS 2024 test set, with or without patient split.}
\end{figure}

We chose to construct the 2024 temporal test set without explicit patient split, because patients do come back to the health system at deployment. While this choice is supported by our clinical collaborators, we did an ablation where the test set excludes seen patients. Keeping the finetuned model fixed and varying the test data to either include or exclude seen patients, we see \textbf{similar test performance on readmission prediction, in hospital mortality prediction, and length of stay prediction}. 
\autoref{fig:patient_control_readmission} shows that on readmission prediction, removing patients seen from pretraining and finetuning slightly increases the performance from  76.5\% (purple bar) to 77\% (grey bar). Similarly, the performance increase is 0.39\% for mortality~(\autoref{fig:patient_control_mortality}) and 0.70\% for LOS~(\autoref{fig:patient_control_los}). 
While surprising, the slight increase could be attributed to \textbf{repeated patients being both older and more likely to be a minority}. For readmission prediction, repeated patients are on average 13 years older than non repeated patients (55 v.s. 42 years). For mortality and LOS prediction, the age gap widens to 16 years (57 v.s. 41 years). In addition, repeated patients include a larger proportion of non-white patients (41\% v.s. 36\%). These findings show that our choice of temporal split does not over-estimate model performance. 

\newpage

\section{External Validation: MIMIC v.s. NYU}\label{app:external_val}

\begin{figure}[htbp]
    \centering
    \begin{subfigure}[t]{0.47\linewidth}
    \includegraphics[width=\linewidth]{figures/main/mimic_validation/mimic_vs_nyu_heatmap.pdf}
    \caption{Readmission prediction.}\label{fig:nyu_vs_mimic_readmission}
    \end{subfigure}\\
     \begin{subfigure}[t]{0.47\linewidth}
        \centering
    \includegraphics[width=\linewidth]{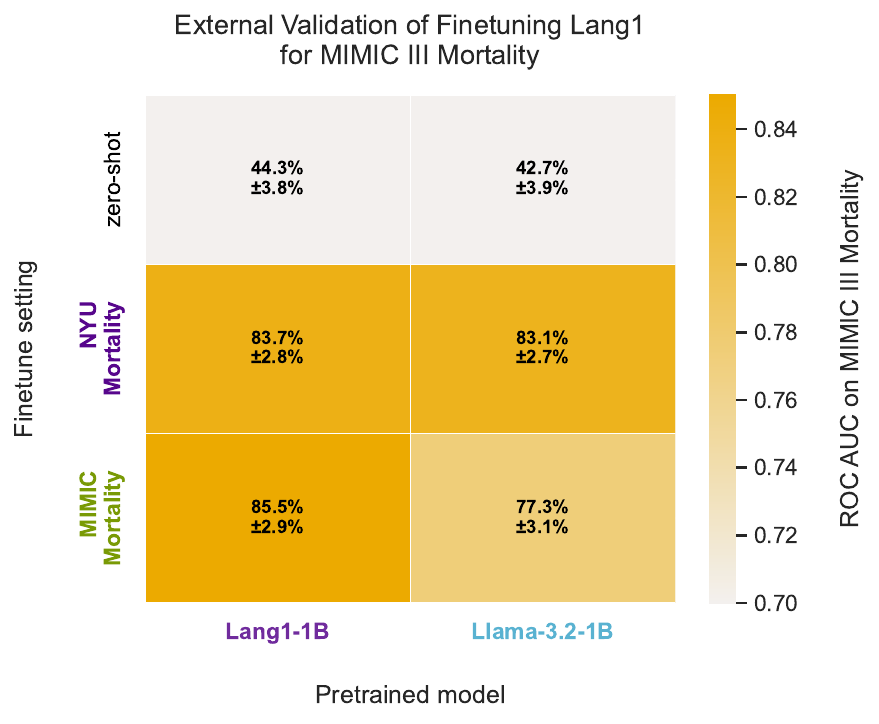}
    \caption{In-hospital Mortality.}
    \label{fig:nyu_vs_mimic_mortality}
     \end{subfigure}
     \hfill
      \begin{subfigure}[t]{0.47\linewidth}
        \centering
    \includegraphics[width=\linewidth]{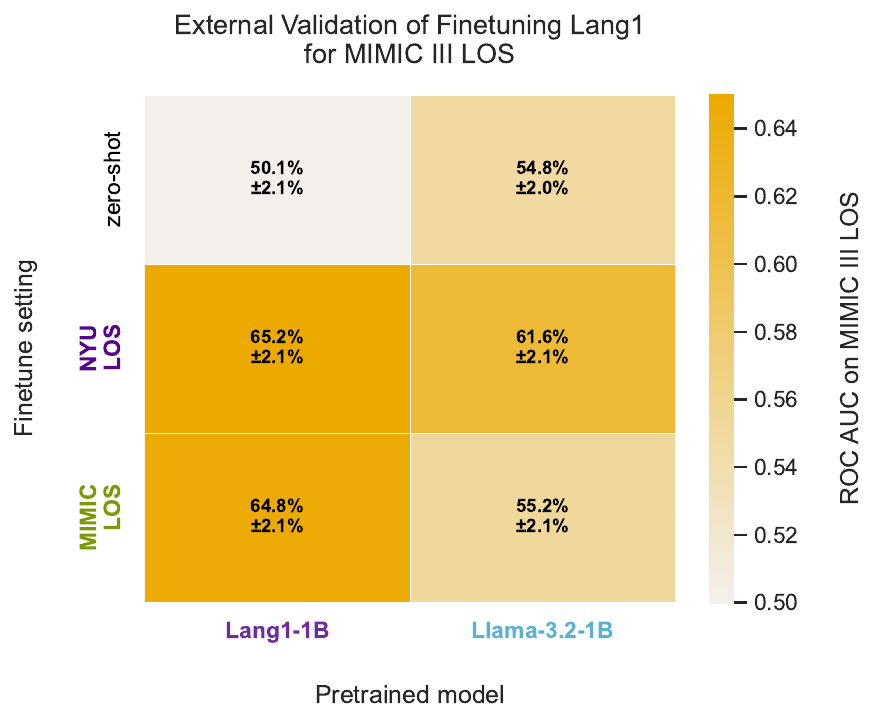}
    \caption{LOS.}
    \label{fig:nyu_vs_mimic_mortality}
     \end{subfigure}
    \caption{Finetuning on NYU transfers well to MIMIC III for readmission, mortality and LOS prediction. Overall performance is better on finetuning \textsc{Lang1-1B} compared to \textsc{Llama 3.2 1B}.}\label{fig:nyu_vs_mimic}
\end{figure}

To check how well \textsc{Lang1} generalizes to a different health system, we compare finetuning \textsc{Lang1-1B} and \textsc{Llama-3.2-1B} on three classification tasks (readmission, mortality, length of stay) from MIMIC III and NYU, and test on MIMIC III. MIMIC III \citep{Johnson2023-zf} is a database of electronic health records from ICU patients at the Beth Israel Hospital in Boston Massachusetts, whereas NYU+ datasets are collected from New York City. See \ref{app:mimic_datasets} for details about MIMIC datasets and Methods \ref{sec:finetune_data} for NYU datasets. 
\autoref{fig:nyu_vs_mimic} shows the heatmaps of MIMIC III test performance for finetuning \textsc{Lang1-1B} and \textsc{Llama-3.2-1B} on NYU or MIMIC-III, with zero-shot performance as baselines. The \textit{x} axis is the pretrained model (purple \textsc{Lang1-1B} and blue \textsc{Llama-3.2-1B}). The \textit{y} axis is the finetuning setting (purple indicates finetuning on NYU data; green indicates finetuning on MIMIC III data; and black indicates zero-shot). Darker yellow cells indicate higher test AUROC on MIMIC III.

\paragraph{Finetuning \textsc{Lang1-1B} on NYU Readmission transfers to MIMIC-III Readmission} The difference in mean AUROC between finetuning \textsc{Lang1-1B} on MIMIC III versus NYU range from 0.4\%-1.8\% AUROC and are roughly within standard error, showing that finetuning \textsc{Lang1-1B} at NYU transfers well to MIMIC. 

\paragraph{\textsc{Lang1-1B} achieves better performance than \textsc{Llama-3.2-1B}} The difference in mean AUROC between finetuning \textsc{Lang1-1B} and \textsc{Llama-3.2-1B} on the same data range from 0.6\%-9.6\%, showing that the clinically pretrained \textsc{Lang1-1B} is a preferred base model for these three clinical predictive tasks.

\begin{figure}[ht!]
    \centering

        \begin{subfigure}[t]{0.43\linewidth}
        \centering
        \includegraphics[width=\linewidth]{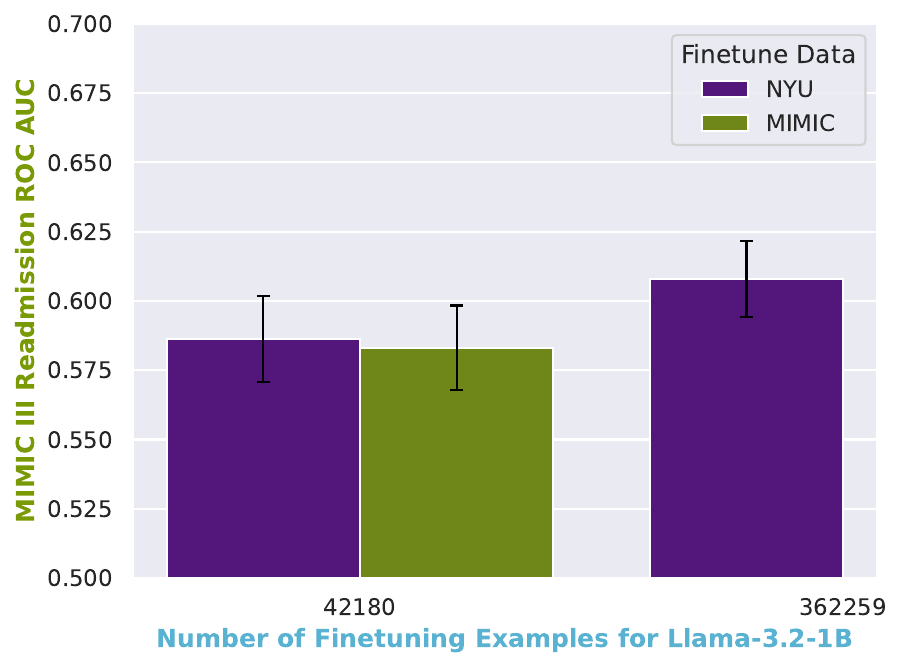}
        \caption{Finetuning on full NYU dataset leads to best performance on \textsc{Llama-3.2-1B}, even though NYU Readmission is not directly in-distribution for MIMIC Readmission.}
        \label{fig:llama_32_eff}
    \end{subfigure}
    \hfill
    \begin{subfigure}[t]{0.43\linewidth}
        \centering
        \includegraphics[width=0.95\linewidth]{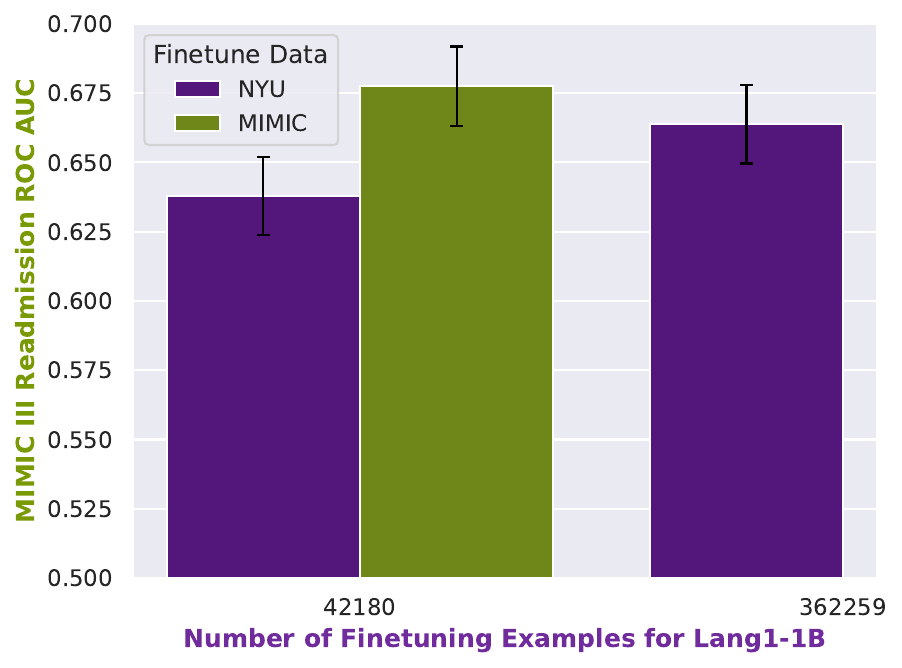}
        \caption{Finetuning on MIMIC~III leads to best performance for \textsc{Lang1-1B}, even though NYU Readmission has more labeled pairs.}
        \label{fig:downsample_nyu_vs_mimic}
    \end{subfigure}

    \caption{Clinical models such as \textsc{Lang1-1B} benefit more from in-domain data, whereas generalist models such as Llama-3.2-1B could benefit from more slightly OOD examples compared to fewer in-distribution examples.}
    \label{fig:mimic_vs_nyu_ablation}
\end{figure}

\paragraph{Finetuning on NYU, which is slightly out-of-distribution, could outperform finetuning on MIMIC for \textsc{Llama-3.2-1B}} It is reasonable to expect that in-distribution finetuning leads to best performance. While this is true for \textsc{Lang1-1B} (purple), it is not the case for \textsc{Llama-3.2-1B} (blue), which sees 2.5\% - 6.4\% AUROC increase from finetuning on NYU compared to finetuning on MIMIC. We hypothesize that this is because NYU data has more labeled pairs, and that \textbf{non clinically pretrained models would benefit more from a larger number of slightly out-of-distribution pairs}. We test this hypothesis on readmission prediction, where NYU Readmission (36,2259 examples) is 8.6 times the size of MIMIC~III readmission (42,180 examples). 
\autoref{fig:llama_32_eff} shows that test performance on MIMIC III is similar when \textsc{Llama-3.2-1B} is finetuned on NYU Readmission that is \textbf{downsampled} to be the same size as MIMIC III (42,180). This pattern does not hold when the pretrained model is changed to the clinically pretrained \textsc{Lang1-1B}: \autoref{fig:downsample_nyu_vs_mimic} shows that finetuning on MIMIC-III readmission yields the best test performance on MIMIC-III, despite fewer number of finetuning examples.


\end{appendices}

\newpage
\bibliography{sn-bibliography}

\end{document}